%% file: paper.tex
\pdfoutput=1

\documentclass{article}

\usepackage[utf8]{inputenc} 
\usepackage[T1]{fontenc}    
\usepackage{hyperref}       
\usepackage{url}            
\usepackage{booktabs}       
\usepackage{amsfonts}       
\usepackage{nicefrac}       
\usepackage{microtype}      
\usepackage{xcolor}         
\usepackage{paper}
\usepackage{adjustbox}

\newtoggle{arxiv}
\toggletrue{arxiv}

\iftoggle{arxiv}
{
\usepackage[numbers]{natbib}
\setlength{\textwidth}{6.5in}
\setlength{\textheight}{9in}
\setlength{\oddsidemargin}{0in}
\setlength{\evensidemargin}{0in}
\setlength{\topmargin}{-0.5in}
\newlength{\defbaselineskip}
\setlength{\defbaselineskip}{\baselineskip}
\setlength{\marginparwidth}{0.8in}
}
{
\usepackage[nonatbib]{neurips_2020}
}

\title{Towards Adversarial Robustness via Transductive Learning}
\date{}
\iftoggle{arxiv}{
\usepackage{authblk}
\author[1]{Jiefeng Chen}
\author[1]{Yang  Guo}
\author[2]{Xi Wu}
\author[1]{Tianqi Li}
\author[3]{Qicheng Lao}
\author[1]{Yingyu Liang}
\author[1]{Somesh Jha}
\affil[1]{University of Wisconsin-Madison\vspace{4pt}}
\affil[2]{Google\vspace{4pt}}
\affil[3]{Mila - Quebec Artificial Intelligence Institute\vspace{4pt}}
\affil[ ]{\texttt{\{jiefeng; yguo; richardl; yliang; jha\}@cs.wisc.edu}, \texttt{wuxi@google.com}, \texttt{qicheng.lao@gmail.com}}
}
{
\usepackage[nonatbib]{neurips_2020}
\author{
}
}

\begin{document}

\maketitle 

\input{abstract}

\section{Introduction}
\label{sec:intro}
\input{intro}

\section{Preliminaries}
\label{sec:preliminaries}
\input{preliminaries}

\section{Modeling Transductive Robustness}
\label{sec:modeling}
\input{modeling}

\section{Adaptive Attacks against Transductive Defenses}
\label{sec:adaptive-attacks}
\input{adaptive-attacks}

\section{New Positive Evidence for the Usefulness of Transductive Learning}
\label{sec:positive-evidence}
\input{positive-evidence}

\section{Experiments}
\label{sec:experiments}
\input{experiments}

{\small
  \bibliographystyle{plain}
  \bibliography{paper}
}

\newpage
\appendix
\input{appendix}

\end{document}

%% file: abstract.tex
\begin{abstract}
  There has been emerging interest to use transductive learning for adversarial robustness
  (Goldwasser et al., NeurIPS 2020; Wu et al., ICML 2020).
  Compared to traditional ``test-time'' defenses,
  these defense mechanisms ``dynamically retrain'' the model based on test time input
  via transductive learning; and theoretically, attacking these defenses boils down to
  bilevel optimization, which seems to raise the difficulty for adaptive attacks.
  In this paper, we first formalize and analyze modeling aspects of transductive robustness.
  Then, we propose \emph{the principle of attacking model space}
  for solving bilevel attack objectives,
  and present an instantiation of the principle which breaks previous transductive defenses.
  These attacks thus point to significant difficulties in the use of
  transductive learning to improve adversarial robustness.
  To this end, we present new theoretical and empirical evidence
  in support of the utility of transductive learning.
\end{abstract}


%% file: intro.tex
Adversarial robustness of deep learning models has received significant attention
in recent years (see the tutorial by~\cite{KM-tutorial} and references therein).
The classic threat model of adversarial robustness considers an {\em inductive setting}
where a model is trained and fixed, and an attacker attempts to thwart the model
with adversarially perturbed input.
This gives rise to a minimax objective for the defender,
and accordingly
adversarial training~\cite{MMSTV18, SND18, schmidt2018adversarially, carmon2019unlabeled}
to improve adversarial robustness.

Going beyond the inductive model, there has been emerging interest in using
\emph{transductive learning} for adversarial robustness
(Goldwasser et al.~\cite{GKKM20}; Wu et al.~\cite{Wu20}).
Basically,  let $U$ denote the clean unlabeled test-time input,
and $U'$ be the (possibly) adversarially perturbed version of $U$,
these work apply a transductive learning algorithm $\Gamma$ to $U'$ to
get an updated model $\Gamma(U')$ to predict on $U'$.
The hope is that this test-time adaptation may be useful to improve adversarial robustness,
because $\Gamma$ is applied \emph{after} the attacker's move of
producing adversarial examples $U'$. This scenario is practically motivated
since many machine learning pipelines are deployed with
batch prediction~\cite{batch-prediction} ($|U'| \gg 1$).

In the first part of this work we formalize a transductive threat model to capture
these defenses. From the attacker perspective, the threat model can be viewed as
considering a transductive optimization objective
$\max_{V' \in N(V)}L_a(\Gamma(F, D, U'), V')$ (Definition~\ref{def:maximin-model},
formula (\ref{obj:trans-attack-original})),
where $L_a$ is a loss function for evaluating the gain of the attack,
$U' = V'|_X$ are the test-time feature vectors, $\Gamma$ is the defender mechanism,
$F$ is a pre-trained model, and $D$ is the labeled training on which $F$ is trained.
This objective is \emph{transductive} because $U'$ appear in both attack (the second
parameter of $L_a$) and defense (input to $\Gamma$).

By choosing different $L_a$ and $\Gamma$,
we show that our threat model captures various defenses,
such as various test-time defenses considered in~\cite{ACW18},
Randomized Smoothing~\cite{CRK19}, as well as~\cite{GKKM20, Wu20}.
We focus the attention of the current work in settings of~\cite{GKKM20, Wu20}
where $\Gamma$ is indeed transductive learning (by contrast,
the $\Gamma$ considered in~\cite{ACW18, CRK19} do not update the model,
but rather sanitize the input).

We then consider principles for adaptive attacks in the transductive model
(principles for the inductive model have been developed in~\cite{Tram20}).
We note that with a transductive learner $\Gamma$, the attacker faces a more
challenging situation: $\Gamma$ is far from being differentiable
(compared to the situations considered in BPDA~\cite{ACW18}),
and the attack set $U'$ also appears in the defense $\Gamma$.
To address these difficulties for adaptive attacks,
our key observation is to consider the \emph{transferability of adversarial examples},
and consider a \emph{robust} version of (\ref{obj:trans-attack-original}):
$\max_{U'}\min_{\overline{U} \in {\cal N}(U')}L_a(\Gamma(\overline{U}), V')$
(formula (\ref{obj:robust-trans-attack})),
where we want to find a \emph{single} attack set $U'$ to thwart a family of models,
induced by $\overline{U}$ ``around'' $U'$. While seemingly more difficult to solve,
this objective relaxes the attacker-defender constraint,
and provides more information in dealing with nondifferentiability.

Based on the principle, we devise new adaptive attacks to thwart previous defenses.
We show that we can break the defense in~\cite{Wu20} in the settings of their consideration.
For~\cite{GKKM20}, we devise attacks against URejectron
in \emph{deep learning settings} and the \emph{small-perturbation regime}
(their theory is developed in the regime of bounded VC-dimensions)
to demonstrate two subtleties:
{\bf (1)} We can generate imperceptible attacks that can
slip through the discriminator trained by URejectron and cause wrong predictions,
and {\bf (2)} We can generate ``benign'' perturbations
(for which the base classifier in URejectron predicts correctly),
but will all be rejected by the discriminator.
Our attacks thus point to significant difficulties in the use of
transductive learning to improve adversarial robustness.

In the final part of the paper,
we give new positive evidence in support of the utility of transductive learning.
We propose Adversarial Training via Representation Matching
(ATRM, (\ref{obj:atrm})) which combines adversarial training with
unsupervised domain adaptation.
ATRM achieves better empirical adversarial robustness
compared to adversarial training alone against our strongest attacks
(on CIFAR-10, we improved from 41.06\% to 53.53\%).
Theoretically, we prove a first separation result
where with transductive learning one can obtain nontrivial robustness,
while without transductive learning, no nontrivial robustness can be achieved.
We also caution the limitations of these preliminary results
(e.g., the high computational cost of ATRM).
Nevertheless, we believe that our results give a step in systematizing the understanding of
the utility of transductive learning for adversarial robustness.



%% file: preliminaries.tex
\textbf{Notations.}
Let $F$ be a model, and for a data point $(\bfx,y) \in \calX \times \calY$,
a loss function $\ell(F;\bfx,y)$ gives the loss of $F$ on the point.
Let $V$ be a set of labeled data points, and let
$L(F, V) = \frac{1}{|V|}\sum_{(\bfx,y) \in V}\ell(F; \bfx, y)$ denote the empirical
loss of $F$ on $V$. For example, if we use binary loss
$\ell^{0,1}(F; \bfx, y) = \mathbbm{1}[F(\bfx) \neq y]$,
this gives the test error of $F$ on $V$.
We use the notation $V|_X$ to denote the projection of $V$ to its features,
that is $\{(\bfx_i, y_i)\}_{i=1}^m|_X \mapsto \{\bfx_i\}_{i=1}^m$.

Throughout the paper, we use $N(\cdot)$ to denote a neighborhood function for
perturbing features: That is, $N(\bfx) = \{ \bfx' \mid  d(\bfx', \bfx)  < \epsilon \}$
is a set of examples that are close to $\bfx$ in terms of a distance metric $d$
(e.g.,  $d(\bfx', \bfx) = \| \bfx'-\bfx \|_p$). Given $U=\{\bfx_i\}_{i=1}^m$,
let $N(U) = \{ \{\bfx'_i\}_{i=1}^m \mid d(\bfx'_i, \bfx_i) < \epsilon, i=0,\dots,m \}$.
Since labels are not changed for adversarial examples, we also use the notation
$N(V)$ to denote perturbations of features, with labels fixed.

\textbf{Dynamic models and transductive learning.}
Goodfellow~\cite{G19} described the concept of {\em dynamic models},
where the model is a moving target that continually changes,
even after it has been deployed. In the setting of transductive learning
for adversarial robustness, this means that after the model is deployed,
every time upon an (unlabeled) input test set $U$ (potentially adversarially perturbed),
the model is updated according to $U$ before producing predictions on $U$.
As is standard for measuring the performance of transducitve learning,
we (only) measure the accuracy of the updated model on $U$.

\textbf{Threat model for classic adversarial robustness.}
The classic adversarial robustness can be written down succinctly as a minimax objective,
$\min_F\Exp_{(\bfx,y)\sim V}\big[\max_{\bfx' \in N(\bfx)}[\ell(F; \bfx', y)]\big]$.
One can reformulate this objective into a game between two players
(for completeness, we record this in Definition~\ref{def:minimax-model} in
Appendix~\ref{sec:minimax-model}). This reformulation (while straightforward) is useful for
formulating more complex threat models below.


%% file: modeling.tex
This section discusses modeling for transductive adversarial robustness.
We first give the formal definition of our threat model,
and show that it encompasses various test-time mechanisms as instantiations.
We use our threat model to analyze in detail the transductive defense
described in~\cite{GKKM20}. We present empirical findings for the subtleties
of Goldwasser et al.'s defense in the small-perturbation regime with typical deep learning.
We end this section by highlighting emphasis of this work.

\subsection{Formulation of the threat model for transductive adversarial robustness}
\label{sec:def-threat-model}

The intuition behind the transductive threat model
is the same as that of the transductive learning~\cite{V98},
except that now the unlabeled data can be adversarially perturbed by an adversary.
Specifically, at test time, after the defender receives
the adversarially perturbed data $U'$ to classify,
the defender trains a model based on $U'$, and the test accuracy is evaluated only for $U'$.
(i.e., for different test set $U$ we may have different models and different test accuracy.)
The formal definition is as follows:

\begin{mdframed}[backgroundcolor=black!10]
  \small
  \begin{definition}[{\bf Transductive threat model for adversarial robustness}]
    \label{def:maximin-model}
    Fix an adversarial perturbation type.
    Let $P_{X, Y}$ be a data generation distribution.
    Attacker is an algorithm $\calA$,
    and defender is a pair of algorithms $(\calT, \Gamma)$,
    where $\calT$ is a supervised learning algorithm,
    and $\Gamma$ is a transductive learning algorithm.
    Let $L_a$ be a loss function to measure the valuation of the attack.\\

    \hrule
    \noindent\centerline{\bf Before the game}\\
    \noindent{\bf Data setup}\\
    $\bullet$\ \  A (clean) training set $D$ is sampled i.i.d.\ from $P_{X, Y}$.\\
    $\bullet$\ \ A (clean) test set $V$ is sampled i.i.d.\ from $P_{X, Y}$.
    \vskip 3pt
    \hrule

    \noindent\centerline{\bf During the game}\\
    \noindent{\bf Training time}\\
    $\bullet$\ \ {\bf (Defender)}
    Train $F = \calT(D)$, using the labeled source data.

    \noindent{\bf Test time}\\
    $\bullet$\ \ {\bf (Attacker)} Attacker receives $V$,
    and produces an (adversarial) unlabeled dataset $U'$ as follows:
    \begin{enumerate}
    \item On input $\Gamma$, $F$, $D$, and $V$, $\calA$ perturbs each point
      $(\bfx,y) \in V$ to $(\bfx', y)$ (subject to the agreed attack type),
      giving $V'=\calA(\Gamma, F, D, V)$ (that is, $V' \in N(V)$).
    \item Send $U' = V'|_X$ (the feature vectors of $V'$) to the defender.
    \end{enumerate}
    $\bullet$ {\bf (Defender)} Produce a model as $F^* = \Gamma(F, D, U')$.
    \vskip 3pt
    \hrule

    \noindent\centerline{\bf After the game}\\
    \noindent{\bf Evaluation (referee)}\\
    The referee computes the valuation $L_a(F^*, V')$.
  \end{definition}
\end{mdframed}

\noindent{\bf White-box attacks}.
An adversary, while cannot directly attack the final model the defender trains,
still has {\em full knowledge of the transductive mechanism $\Gamma$} of the defender,
and can leverage that for adaptive attacks.
Note that, however, the adversary does not know the private randomness of $\Gamma$.

\textbf{Examples}. The threat model is general to encompass various defenses.
We give a few examples (some may not be entirely obvious).
\begin{example}[\textbf{Test-time defenses}]
  There have been numerous proposals of ``test-time defenses'' for adversarial robustness.
  These defenses can be captured by a pair $({\cal T}, \Gamma)$ where ${\cal T}$
  trains a fixed pretrained model $F$, and $\Gamma$ is a ``non-differentiable'' function
  which sanitizes the input and then send it to $\Gamma$.
  Most of these proposals were broken by BPDA~\cite{ACW18}.
  We note that, however, these proposals are far from transductive learning:
  There is no transductive learning that trains the model using the test inputs,
  and very often these algorithms even applied to single test points (i.e. $|U|= 1$).
\end{example}

\begin{example}[\textbf{Randomized smoothing}~\cite{CRK19}]
  \label{example:randomized-smoothing}
  Another interesting proposal that falls under our modeling is randomized smoothing.
  We describe the construction for $|U|=1$, and it is straightforward to extend it to
  $|U|>1$. $\cal T$ prepares a fixed pretrained model $F$. $\Gamma$ works as follows:
  Upon a test feature $\bfx'$, $\Gamma$ samples a random string
  $\xi = (\varepsilon_1, \dots \varepsilon_n)$,
  consisting of $n$ independent random noises.
  Then $\Gamma$ returns the prediction function $\predict[\xi]$
  ($\predict$ is described in ``{\bf Pseudocode for certification and prediction}''
  on top of page 5,~\cite{CRK19}), which is $\predict$ with randomness fixed.
  It is straightforward to check that this construction is equivalent to using
  $\predict$ at the test time where noises are sampled internally
  (we simply move sampling of noises outside into $\Gamma$,
  and return $\predict[\xi]$ as a model).
  This is an important example for the utility of private randomness.
  In fact, if $n$ is small, then an adversary can easily fail the defense
  for any fixed random noise.
\end{example}

Neither of the two examples above uses a $\Gamma$ that does ``learning'' on the
unlabeled data. Below we analyze examples where $\Gamma$ indeed does transductive learning.
\begin{example}[\textbf{Runtime masking and cleansing}]
  Runtime masking and cleansing~\cite{Wu20} (RMC) is a recent proposal that uses test-time
  learning (on the unlabeled data) to enhance adversarial robustness.
  The $\Gamma$ works with $|U|=1$, and roughly speaking,
  updates the model by solving 
  $F^* = \argmin_F \sum_{(\bfx, y) \in N'(\widehat{\bfx})}L(F, \bfx, y)$,
  where $\widehat{\bfx}$ is the test time feature point.
  In this work, we develop strong adaptive attacks to break this defense.
\end{example}

\begin{example}[\textbf{Unsupervised Domain Adaptation (as transductive learning)}]
  While not explored in previous work, our modeling also indicates a natural application
  of unsupervised domain adaptation, such as $\dann$~\cite{ajakan2014domain},
  as \emph{transductive learning} for adversarial robustness:
  Given $U'$, we train a DANN model on the training dataset $D$ and $U$,
  and then evaluate the model on $U'$. In our experiments,
  we show that this alone already provides better robustness than RMC,
  even though we can still thwart this defense using our strongest adaptive attacks.
\end{example}

\subsection{Goldwasser et al.'s transductive threat model and URejectron}
\label{sec:goldwasser}
While seemingly our formulation of the transductive threat model is quite different
from the one described in~\cite{GKKM20}, in fact one can recover their threat model
naturally (specifically, the transductive guarantee described in Section 4.2 of their paper):
First, for the perturbation type,
we simply allow arbitrary perturbations in the threat model setup.
Second, we have a fixed pretrained model $F$, and the adaptation algorithm $\Gamma$
learns a set $S$ which represents the set of ``allowable'' points
(so $F|_S$ gives the predictor with redaction, namely it outputs $\perp$ for points
outside of $S$). Third, we define two error functions as (5) and (6) in~\cite{GKKM20}:
\begin{align}
  \small
  &\err_{U'}(F|_S, f) \equiv \frac{1}{|U'|}\bigg|\bigg\{
  \bfx' \in U' \cap S\bigg| F(\bfx')\neq f(\bfx') \bigg\}\bigg|\\
  &\rej_{U}(S) \equiv \frac{|U \setminus S|}{|U|}
\end{align}
where $f$ is the ground truth hypothesis.
The first equation measures prediction errors in $U'$ that passed through $S$,
and the second equation measures the rejection rate of the clean input.
Finally, we define the following loss function for valuation,
which measures the two errors as a pair:
\begin{align}
  \label{formula:GKKM-trans-guarantee}
  \small
  L_a(F|_S, V') = \left( \err_{U'}(F|_S), \rej_U(S) \right)
\end{align}
The theory in~\cite{GKKM20} is developed in the bounded VC dimension scenarios.
Specifically, Theorem 5.3 of their paper (Transductive Guarantees) establishes for
$\Gamma=\rejectron$ that, for proper $\varepsilon^*$, with high probability over
$D, V \sim P_{X,Y}^n$, $L_a(\rejectron(D, U'), V') \le (\varepsilon^*, \varepsilon^*)$,
for \emph{any} $U'$.

\textbf{Subtleties of URejectron}.
\cite{GKKM20} also derived an unsupervised version, $\urejectron$, of $\rejectron$,
with similar theoretical guarantees, and presented corresponding empirical results.
Based on their implementation, we studied $\urejectron$ in the setting of
\emph{deep learning with small perturbations}.
Specifically, we evaluated URejectron on GTSRB dataset using ResNet18 network.
The results are shown in Figure~\ref{fig:urejectron-exp}.
Figure~\ref{fig:urejectron-exp}(a) shows that for \emph{transfer attacks} generated
by PGD attack~\cite{MMSTV18}, URejectron can indeed work as expected.
However, by using different attack algorithms, such as CW attacks~\cite{CW17},
(nevertheless these attacks are \emph{transfer attacks},
which are \emph{weak} instantiations of
our framework described in Section~\ref{sec:adaptive-attacks}),
we observe two possible failure modes:

\begin{figure*}[t]
  \centering
  \begin{subfigure}{0.3\linewidth}
    \centering
    \includegraphics[bb=0 0 260 225, width=\linewidth]{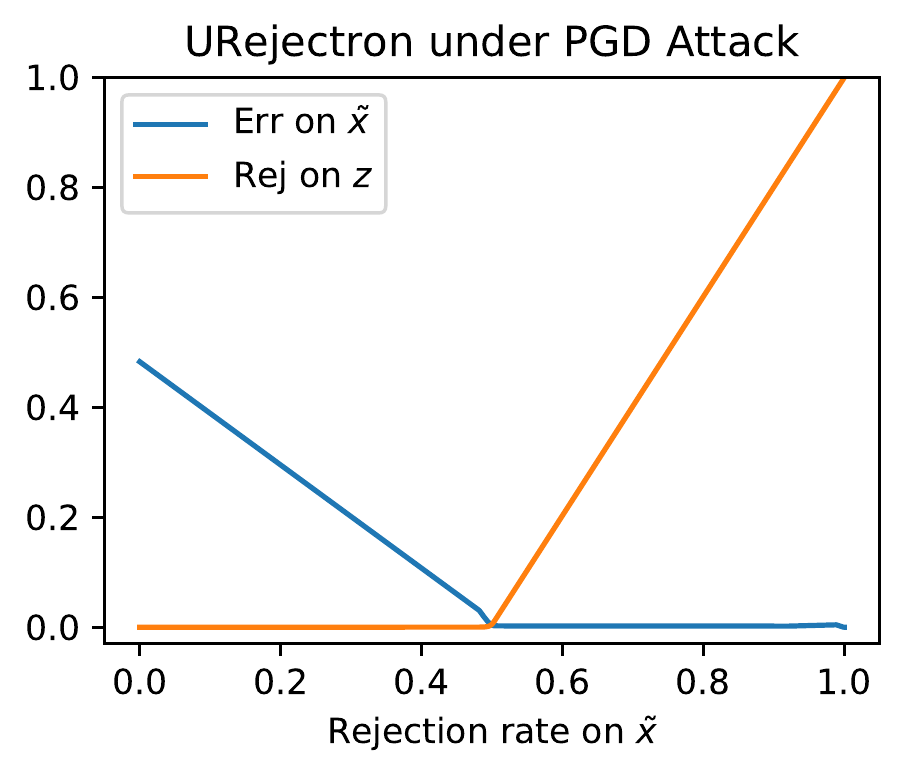}
    \caption{}
  \end{subfigure}
  \begin{subfigure}{0.3\linewidth}
    \centering
    \includegraphics[bb=0 0 260 225,width=\linewidth]{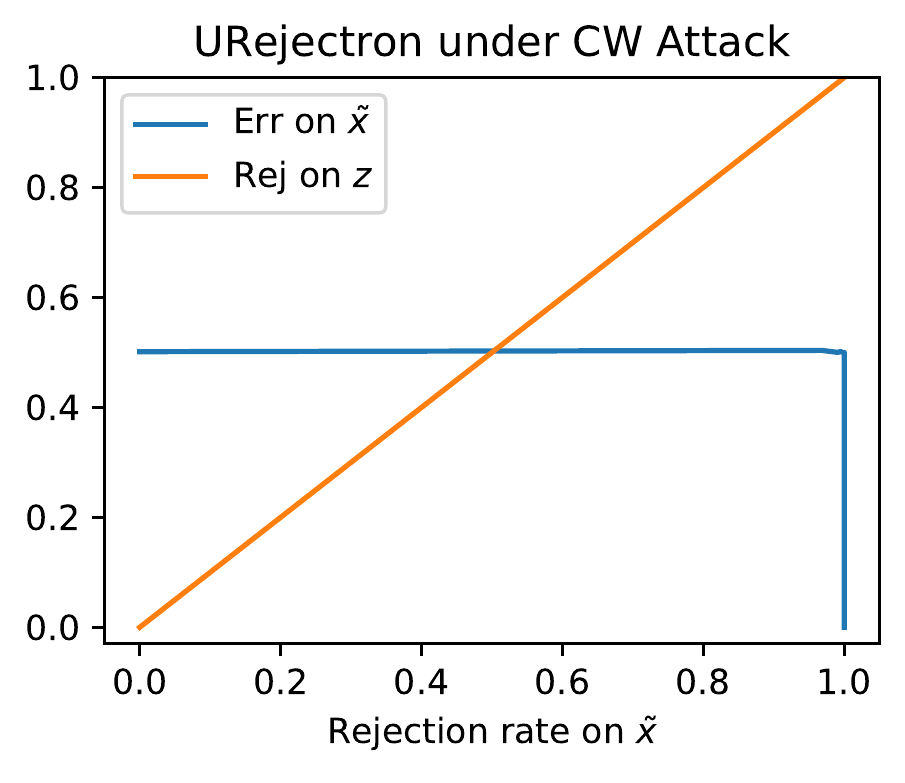}
    \caption{}
  \end{subfigure} 
  \begin{subfigure}{0.3\linewidth}
    \centering
    \includegraphics[bb=0 0 260 225,width=\linewidth]{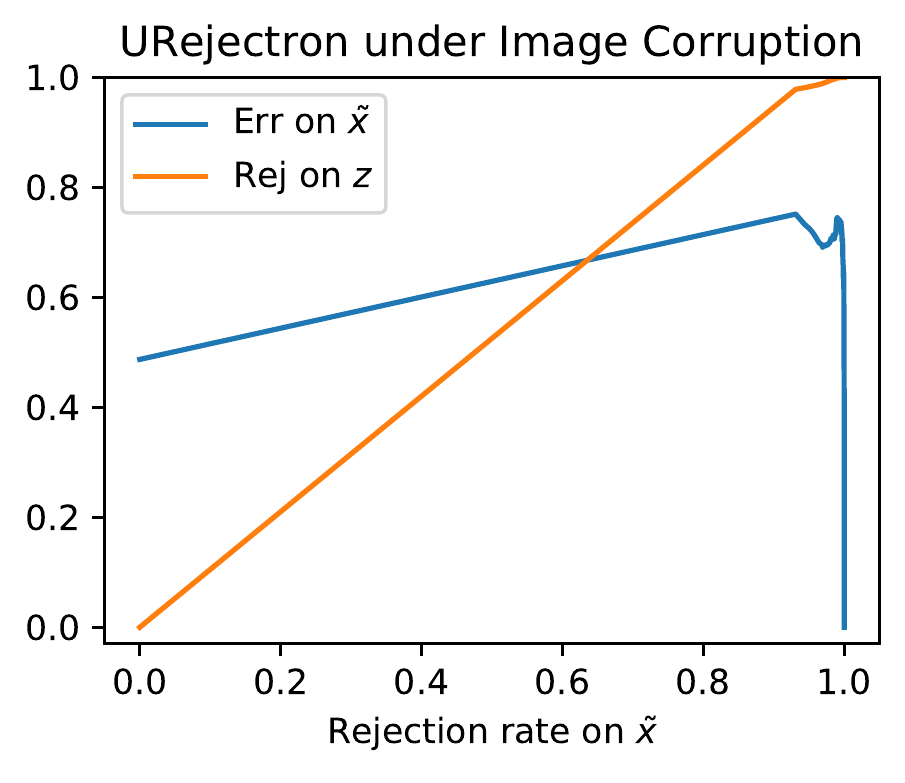}
    \caption{}
  \end{subfigure} 
  \caption{\small URejectron in three settings.
    $z$ contains ``normal'' examples on which the classifier can have high accuracy.
    $\tilde{x}$ includes $z$ and consists of a mix of 50\% ``normal'' examples
    and 50\% adversarial examples.
    In (a), the normal examples are clean test inputs
    and the adversarial examples are generated by PGD attack~\cite{MMSTV18}.
    In (b), the ``normal'' examples are still clean test inputs
    but adversarial examples are generated by CW attack~\cite{CW17}.
    In (c), the ``normal'' examples are generated by corruptions~\cite{Hendrycks19}
    (adversarial examples are generated by PGD attacks). 
  }
  \label{fig:urejectron-exp}
\end{figure*}

\textbf{\em Imperceptible adversarial perturbations that slip through}.
Figure~\ref{fig:urejectron-exp}(b) shows that one can construct adversarial examples that are very similar to
the clean test inputs that can slip through their URejectron construction of $S$
(in the deep learning setting), and cause large errors.

\textbf{\em Benign perturbations that get rejected}.
Figure~\ref{fig:urejectron-exp}(c) shows that one can generate ``benign'' perturbed examples
(i.e., the base classifier can give correct predictions),
using image corruptions such as slightly increasing the brightness,
but URejectron rejects them all. While strictly speaking,
this failure mode is beyond their guarantee (\ref{formula:GKKM-trans-guarantee}),
this indicates that in the small-perturbation regime
things can be more subtle compared to the seemingly harder ``arbitrary perturbation'' case.

We thus believe that a more careful consideration
for the small-perturbation regime is warranted.

\subsection{Focus of the current work}
While we have shown that our transductive threat model formulation is quite encompassing,
in this work we focus on a regime that differ from the considerations in~\cite{GKKM20}:

\noindent\textbf{\em Small perturbations}.
Arbitrary perturbations include small perturbations as a special case.
In this work, we are primarily motivated to study situations
where test and training samples are connected,
instead of arbitrarily far away. For this reason,
we focus on the \emph{small-perturbation regime}.

\noindent\textbf{\em Deep learning with no redaction.}
Due to the previous consideration,
and our main motivation to study the utility of transductive learning,
we focus on the case with no redaction. As for practicality considerations,
we focus on deep learning, instead of learners with bounded VC dimensions.

Finally, different from test-time defenses and randomized smoothing, in this work
we focus on $\Gamma$ that perform actual transductive learning (i.e. update the model
based on unlabeled data).


%% file: adaptive-attacks.tex
In this section we consider adaptive attacks against transductive defenses
under the \emph{white-box} assumption: The attacker knows all the details of the
defender transductive learning algorithm $\Gamma$ (except private randomness used
by the defender). We deduce a principle for adaptive attacks,
which we call \emph{the principle of attacking model space}:
It suggests that effective attacks against a transductive defense may need to consider
\emph{attacking a small set of representative models}.
We give concrete instantiations of this principle, and show in experiments that
they break previous transductive defenses, and is much stronger than attacks
directly adapted from literature on solving bilevel optimization
objectives in deep learning.

\subsection{Goal of the attacker and challenges}
To start with, given a defense mechanism $\Gamma$,
the objective of the attacker can be formulated as:
\begin{align}
  \label{obj:trans-attack-original}
  \max_{V' \in N(V), U'=V'|_X} L_a(\Gamma(F, D, U'), V').
\end{align}
where $L_a$ is the loss function of the attacker. We make some notational simplifications:
Since $D$ is a constant, in the following we drop it and write $\Gamma(U')$.
Also, since the attacker does not modify the labels in the threat model,
we abuse the notation (one can think as hard-wiring labels into $L_a$),
and write the objective as
\begin{align}
  \label{obj:trans-attack}
  \max_{V',U'=V'|_X} L_a(\Gamma(U'), U').
\end{align}
A generic attacker would proceed iteratively as follows:
It starts with the clean test set $V$, and generates a sequence of
\emph{(hopefully) increasingly stronger} attack sets
$U^{(0)} = V|_X, U^{(1)}, \dots, U^{(i)}$.
We note several basic but important {\em differences} between transductive attacks
and inductive attacks in the classic minimax threat model:

\textbf{(D1) $\Gamma(U')$ is \emph{not} differentiable}.
For the scenarios we are interested in, $\Gamma$ is an optimization algorithm
to solve an objective $F^* \in \argmin_{F}L_d(F, D, U')$. This renders
(\ref{obj:trans-attack}) into a bilevel optimization problem~\cite{Colson07anoverview}:
\begin{align}
  \label{formula:attack-obj}
  \begin{split}
    &\max_{V' \in N(V); U'=V'\mid_X} L_a(F^*, V') \\
    &\text{subject to: } F^* \in
    \argmin_{F}L_d(F, D, U'),
  \end{split}
\end{align}
In these cases, $\Gamma$ is in general \emph{not} (in fact far from) differentiable.
A natural attempt is to approximate $\Gamma$ with a differentiable function,
using theories such as Neural Tangent Kernels~\cite{NEURIPS2018_5a4be1fa}.
Unfortunately no existing theory applies to the transductive training,
which deals with unlabeled data $U'$ (also, as we have remarked previously,
tricks such as BPDA~\cite{ACW18} also does not apply because transductive learning
is much more complex than test-time defenses considered there).

\textbf{(D2) $U'$ appears in \emph{both} attack and defense}.
Another significant difference is that the attack set $U'$ also appears as
the input for the defense (i.e. $\Gamma(U')$).
Therefore, while it is easy to find $U'$ to fail $\Gamma(\overline{U})$
for any fixed $\overline{U}$,
it is much harder to find a \emph{good direction} to update the attack and converge to
\emph{an attack set $U^*$ that fails an entire model space induced by itself: $\Gamma(U^*)$}.

\textbf{(D3) $\Gamma(U')$ can be a {\em random variable}}.
In the classic minimax threat model, the attacker faces a fixed model.
However, the output of $\Gamma$ can be a \emph{random variable of models}
due to its private randomness, such as the case of Randomized Smoothing
(Example~\ref{example:randomized-smoothing}). In these cases, successfully attacking
a single sample of this random variable does not suffice.

\begin{algorithm*}[htb]
  \small
  \caption{\textsc{\small Fixed Point Attack ($\fpa$)}}
  \begin{algorithmic}[1]
    \REQUIRE A transductive learning algorithm $\Gamma$, an optional training dataset $D$,
    a natural test dataset $V$, an initial model $F^{(0)}$,
    and an integer parameter $T \ge 0$ (the number of iterations).
    \FOR{$i=0,1,\dots,T$}
    \STATE Attack the model obtained in the last iteration to get the perturbed set: 
    \begin{align}
      \label{obj:gmsa}
      V^{(i)} = \argmax_{V' \in N(V)} L_a(F^{(i)}, V')
    \end{align}
    where $L_a$ is the standard test loss. Set $U^{(i)}=V^{(i)}\mid_X$.
    \STATE Run the transductive learning algorithm $\Gamma$ to get the next model: 
    $F^{(i+1)} =\Gamma(D, U^{(i)})$.
    \ENDFOR
    \STATE Select the best attack set $U^{(k)}$ as
    $k = \argmax_{0 \leq i \leq T} L_a(F^{(i+1)}, V^{(i)})$.
    \RETURN $U^{(k)}$.
  \end{algorithmic}
  \label{alg:fpa}
\end{algorithm*}
\textbf{Fixed Point Attack: A first attempt}.
We adapt previous literature for solving bilevel optimization
in deep learning setting~\cite{journals/corr/abs-1802-09419}
(designed for supervised learning). The idea is simple:
At iteration $i+1$, we fix $U^{(i)}$ and model space $F^{(i)} = \Gamma(U^{(i)})$,
and construct $U^{(i+1)}$ to fail it. We call this the Fixed Point Attack ($\fpa$),
as one hopes that this process converges to a good fixed point $U^*$.
Unfortunately, we found $\fpa$ to be weak in experiments.
The reason is exactly {\bf (D2)}:
$U^{(i+1)}$ failing $F^{(i)}$ may not give any indication that
it can also fail $F^{(i+1)}$ induced by itself.

\subsection{Strong adaptive attacks from attacking model spaces}
\label{sec:strong-attacks}
To develop stronger adaptive attacks, we consider a key property of the
adversarial attacks: The \emph{transferability of adversarial examples}.
Various previous work have identified that
adversarial examples transfer~\cite{46153,DBLP:journals/corr/LiuCLS16},
even across vastly different architectures and models.
Therefore, if $U'$ is a good attack set, we would expect that $U'$
also fails $\Gamma(\overline{U})$ for $\overline{U}$ close to $U'$.
This leads to the consideration of the following objective:
\begin{align}
  \label{obj:robust-trans-attack}
  \max_{U'} \min_{\overline{U} \in {\cal N}(U')} L_a(\Gamma(\overline{U}), U').
\end{align}
where ${\cal N}(\cdot)$ is a neighborhood function (possibly different than $N$).
It induces a family of models $\{\Gamma(U')\ |\ U' \in {\cal N}(U^*)\}$,
which we call a \emph{model space}.
(in fact, this can be a family of random variables of models)
This can be viewed as a natural \emph{robust} version of (\ref{obj:trans-attack})
by considering the transferability of $U'$.
While this is seemingly even harder to solve, it has several benefits:

\textbf{(1) Considering a model space naturally strengthens $\fpa$.}
$\fpa$ naturally falls into this formulation as a weak instantiation where we consider
a single $\overline{U}=U^{(i)}$. Also, considering a model space gives the attacker
more information in dealing with the nondifferentiability of $\Gamma$ {\bf (D1)}.

\textbf{(2) It relaxes the attacker-defender constraint (D2).}
Perhaps more importantly, for the robust objective,
we no longer need the same $U'$ to appear in both defender and attacker.
Therefore it gives a natural relaxation which makes attack algorithm design easier.

In summary, while ``brittle'' $U'$ that does not transfer may indeed exist theoretically,
their identification can be challenging algorithmically, and its robust variant
provides a natural relaxation considering both algorithmic feasibility and attack strength.
This thus leads us to the following principle:

\begin{mdframed}[backgroundcolor=black!10]
  \small
  \textbf{The Principle of Attacking Model Spaces}.
  \emph{An efficient and effective adaptive attack strategy against a transductive defense
    may need to consider \emph{a model space} induced by different $\overline{U}$'s,
    and then identify an attack set to fail all of them.}
\end{mdframed}
\vskip -5pt
\begin{algorithm*}[htb]
  \small
  \caption{\textsc{\small Greedy Model Space Attack (GMSA)}}
  \begin{algorithmic}[1]
    \REQUIRE A transductive learning algorithm $\Gamma$, an optional training dataset $D$,
    a natural test dataset $V$, an initial model $F^{(0)}$,
    and an integer parameter $T \ge 0$ (the number of iterations).
    \FOR{$i=0,1,\dots,T$}
    \STATE Attack the previous models to get the perturbed set: 
    \begin{align}
      \label{obj:gmsa}
      V^{(i)} = \argmax_{V' \in N(V)} L_{\textrm{GMSA}}(\{F^{(j)} \}_{j=0}^i, V')
    \end{align}
    where $L_{\textrm{GMSA}}$ is a loss function. Set $U^{(i)}=V^{(i)}\mid_X$.
    \STATE Run the transductive learning algorithm $\Gamma$ to get the next model: 
    $F^{(i+1)} =\Gamma(D, U^{(i)})$.
    \ENDFOR
    \STATE Select the best attack $U^{(k)}$ as
    $k = \argmax_{0 \leq i \leq T} L_a(F^{(i+1)}, V^{(i)})$,
    \RETURN $U^{(k)}$.
  \end{algorithmic}
  \label{alg:ensemble-attack}
\end{algorithm*}
\textbf{An instantiation: Greedy Model Space Attack (GMSA)}.
We give a simplest possible instantiation of the principle,
which we call the \emph{Greedy Model Space Attack} (Algorithm~\ref{alg:ensemble-attack}).
In experiments we use this instantiation to break previous defenses.
In this instantiation, the family of model spaces to consider is just all the model
spaces constructed in previous iterations (line 2).
$L_{\text{GMSA}}$ (\ref{obj:gmsa})is a loss function that the attacker uses to
attack the history model spaces. We consider two instantiations:
(1) $L^{\text{AVG}}_{\text{GMSA}}(\{F^{(j)} \}_{j=0}^i, V')
= \frac{1}{i+1} \sum_{j=0}^i L(F^{(j)}, V')$,
(2) $L^{\text{MIN}}_{\textrm{GMSA}}(\{F^{(j)}\}_{j=0}^i, V')
= \min_{0 \leq j \leq i} L(F^{(j)}, V')$,
where $L^{\text{AVG}}_{\text{GMSA}}$ gives attack algorithm GMSA-AVG,
and $L^{\text{MIN}}_{\text{GMSA}}$ gives attack algorithm GMSA-MIN.
We solve (\ref{obj:gmsa}) via Projected Gradient Decent (PGD)
(the implementation details of GMSA can be found in Appendix~\ref{sec:gmsa-implementation-detail}).


%% file: positive-evidence.tex
In the experiments we will show that the new adaptive attacks devised
in this paper breaks previous defenses (in typical deep learning settings).
This thus points to significant difficulties in the use of
transductive learning to improve adversarial robustness.
To this end, we provide new positive evidence:
Empirically, we show that by combining adversarial training
and unsupervised domain adaptation (ATRM),
one can indeed obtain improved adversarial robustness compared to adversarial training alone,
against our strongest attacks. Theoretically, we prove a separation result
which demonstrates the utility of transductive learning.
We caution the limitations of these preliminary results,
such as the high computational cost of ATRM. Nevertheless,
these represent a step to systematize the understanding of the utility of
transductive learning for adversarial robustness.

\subsection{Adversarial Training via Representation Matching (ATRM)}
We consider a transductive learning version of adversarial training
where we not only perform adversarial training but also align the representations
of the adversarial training examples and the given test inputs.
Specifically, we consider models $F(\bfx)=c(\phi(\bfx))$
that is a composition of a prediction function $c$ and a representation function $\phi$.
We propose to train the dynamic model with the following objective:
\begin{align}
  \label{obj:atrm}
  \begin{split}
    \minimize_{F} & \mathbb{E}_{(\bfx',y)\in D' } [
    \ell(F(\bfx'),y)] + \alpha \cdot d(p^\phi_{D'}, p^\phi_{U'}) \\
    \text{where } D' & = \argmax_{D' \in N(D)}
    \mathbb{E}_{(\bfx',y)\in D' } \ell(F(\bfx'),y)
  \end{split}
\end{align}
where $N(D)$ is a collection of perturbed sets of $D$ and
$d(p^\phi_{D'}, p^\phi_{U'})$ is the distance between the distribution of
$\phi(\bfx)$ on $D'$ and that on $U'$. This method,
named as Adversarial Training via Representation Matching (ATRM),
achieves positive results in our experiments in Section~\ref{sec:experiments}. 

\subsection{Transductive vs. Inductive: A Separation Result}
We now turn to new theoretical evidence about the usefulness of transductive learning.
A basic theoretical question is \emph{
  whether the transductive setting allows better defense than
  the traditional inductive setting.}
We observe that, by the max–min inequality,
the defender's game value in the former is \emph{no worse}
than that in the latter (see Appendix~\ref{sec:valuation} for proofs).
While this conclusion doesn't involve algorithms,
we also observe that the same holds when algorithms are considered:
Any defense algorithm in the inductive setting can be used as an adaptation algorithm
$\Gamma$ in the transductive setting (which simply ignores the test inputs and outputs
the model trained on the training data), and obtains \emph{no worse} results.
However, these observations only imply that the transductive defense is no harder,
but does not imply that it can be strictly better.

\textbf{A separation result.}
We thus consider a further question:
\emph{Are there problem instances for which there exist defense algorithms
  in the transductive setting with strictly better performance than any algorithm
  in the inductive setting?}
Here a problem instance is specified by a family of data distributions $P_{X, Y}$,
the feasible set of the adversarial perturbations, the number of training data points,
and the number of test inputs for the transductive setting.
We answer this positively by constructing such problem instances
(see Appendix~\ref{sec:gaussian} for proofs):
\begin{theorem}
  \label{thm:gaussian}
  For any $\epsilon \in (0, \frac{1}{3})$, there exist problem instances of
  binary classification with 0-1 loss:
  \begin{itemize}
  \item[(1)] In the inductive threat setting,
    the learned model by {\em any} algorithms must have a large loss
    at least $\frac{1}{2} (1- \epsilon)$.
  \item[(2)] In the transductive threat setting,
    there exist polynomial-time algorithms $\mathcal{T}$ and $\Gamma$ such that 
    the adapted model has a small loss at most $\epsilon$. 
  \end{itemize}
\end{theorem}
It is an intriguing direction to generalize this result to a broader class of problems.


%% file: experiments.tex
This section evaluates several transductive-learning based defenses.
Our findings are summarized as follows
(Appendix~\ref{sec:exp-details} gives details for replicating results):
{\bf (1)} Using Fixed Point Attack ($\fpa$), one can already thwart RMC~\cite{Wu20}.
{\bf (2)} For transductive defenses that are robust to $\fpa$
(e.g. RMC$^+$ and DANN), the GMSA can thwart them.
{\bf (3)} Our ATRM defense provides significant improvement in adversarial robustness,
compared to the adversarial training alone, against our strongest attacks.
For all experiments, the defender uses his own private randomness,
which is different from the one used by the attacker.
Without specified otherwise, all reported values are percentages.

\subsection{Attacking Runtime Masking and Cleansing Defense}

\textbf{Runtime masking and cleansing (RMC)~\cite{Wu20}}.
RMC claimed to achieve state-of-the-art robustness under several adversarial attacks.
However, those attacks are not adaptive attacks since
the attacker is unaware of the defense mechanism.
We thus evaluate RMC with our adaptive attacks.
We assume that the attacker can simulate the adaptation process to
generate a sequence of adversarial examples for evaluation. 
The results are in Table~\ref{tab:rmc-adaptive-attack-exp}:
RMC with the standard model is already broken by FPA attack (which is weaker than GSMA).
Compared to the defense-unaware PGD attack, our  GMSA-AVG attack reduces the robustness from
$98.30\%$ to $0.50\%$ on MNIST and from $97.60\%$ to $8.00\%$ on CIFAR-10.
Further, RMC with adversarially trained model actually provides
{\em worse} adversarial robustness than using adversarial training alone.
Under our GMSA-MIN attack, the robustness is reduced from $96.10\%$ to $58.80\%$ on MNIST
and from $71.70\%$ to $39.60\%$ on CIFAR-10.

\begin{table*}[t!]
  \small
  \centering
  \begin{tabular}{l|l|c|cccc}
    \toprule
    \multirow{2}{0.08\linewidth}{\textbf{Dataset}} & \multirow{2}{0.1\linewidth}{\textbf{Model}} & \multirow{2}{0.1\linewidth}{{{\bf Accuracy}}} &  \multicolumn{4}{c}{\textbf{Robustness}}  \\ \cline{4-7}
                                                   &  &  & PGD & FPA & GMSA-AVG & GMSA-MIN \\  \hline \hline 
    \multirow{2}{0.08\linewidth}{{{\bf MNIST}}}  
                                                   & Standard  & 99.00 & 98.30 & 0.60 & 0.50 & 1.10 \\  
                                                   & Madry et al. & 97.00 & 96.10 & 59.50 & 61.40 & 58.80 \\ \hline
    \multirow{2}{0.08\linewidth}{{{\bf CIFAR-10}}}  
                                                   & Standard  & 93.10 & 97.60 & 8.50 & 8.00 & 8.10 \\  
                                                   & Madry et al. & 90.90 & 71.70 & 40.80 & 42.50 & 39.60 \\  
    \bottomrule
  \end{tabular}
  \caption[]{\small Results on RMC.
    The PGD attack the one used in~\cite{Wu20}.}
  \label{tab:rmc-adaptive-attack-exp}
\end{table*}

\textbf{RMC and RMC$^+$ (an extension of RMC) under the PGD-skip
  attack setting proposed in~\cite{Wu20}}.
In this setting, the attacker generates an adversarial example $\hat{\bfx}^{(p+1)}$
against the network that has been adapted to $\hat{\bfx}^{(1)},\cdots, \hat{\bfx}^{(p)}$.
(we did not consider the weakened versions of PGD-skip attack,
such as PGD-Skip-Partial and PGD-Skip-Delayed attacks,
since they are weaker than PGD-skip and they actually limit the attacker's power
in the white-box setting). The results are in Table~\ref{tab:rmc-pgd-skip-exp}:
As observed in~\cite{Wu20}, under PGD-skip attack, RMC provides limited robustness,
while RMC$^+$ achieves good robustness.
Also, we find that RMC$^+$ is also somewhat robust under the FPA attack.
However, our GMSA-MIN attack (which is stronger than both PGD-skip and FPA attacks),
breaks both RMC and RMC$^+$.
For example, on MNIST, GMSA-MIN attack reduces the robustness of RMC$^+$
with adversarially trained model from $69.80\%$ to $4.00\%$. On
CIFAR-10, the robustness of RMC$^+$ with standard model is reduced from
$60.70\%$ to $4.80\%$ and the robustness of RMC$^+$ with adversarially trained model
is reduced from $71.70\%$ to $23.50\%$ (which is worse than adversarial training alone).

\begin{table*}[t!]
  \small
  \centering
  \begin{tabular}{l|l|l|cccc}
    \toprule
    \multirow{2}{0.08\linewidth}{\textbf{Dataset}} & \multirow{2}{0.12\linewidth}{\textbf{Model}}  &  \multirow{2}{0.08\linewidth}{\textbf{Method}} & \multicolumn{4}{c}{\textbf{Robustness}}  \\ \cline{4-7}
                                                   &  &  & PGD-skip & FPA & GMSA-AVG & GMSA-MIN \\  \hline \hline
    \multirow{4}{0.08\linewidth}{{{\bf MNIST}}}  
                                                   & \multirow{2}{0.12\linewidth}{Standard} & RMC & 0.70 & 0.10 & 0.00 & 0.00 \\  
                                                   &  & RMC$^+$ & 89.60 & 80.70 & 27.10 & 3.40 \\ \cline{2-7}
                                                   & \multirow{2}{0.12\linewidth}{Madry et al.} & RMC & 6.40 & 3.70 & 0.40 & 0.10 \\ 
                                                   &  & RMC$^+$ & 85.50 & 69.80 & 9.60 & 4.00 \\ \hline
    \multirow{4}{0.08\linewidth}{{{\bf CIFAR-10}}}  
                                                   & \multirow{2}{0.12\linewidth}{Standard} & RMC & 0.40 & 0.10 & 0.00 & 0.00 \\  
                                                   &  & RMC$^+$ & 75.10 & 60.70 & 17.60 & 4.80 \\ \cline{2-7}
                                                   & \multirow{2}{0.12\linewidth}{Madry et al.} & RMC & 34.80 & 33.10 & 21.60 & 16.80 \\ 
                                                   &  & RMC$^+$ & 81.50 & 71.70 & 37.30 & 23.50 \\
    \bottomrule
  \end{tabular}
  \caption[]{\small Results on RMC and RMC$^+$ under the PGD-skip attack setting.
    We set $p=100$. }
  \label{tab:rmc-pgd-skip-exp}
\end{table*}

\subsection{Unsupervised Domain Adaptation and ATRM }
\textbf{Attacking DANN}.
We also evaluate DANN (alone) as a transductive learning mechanism for
adversarial robustness. The results are presented in Table~\ref{tab:dann-atrm-eval-exp}.
Interestingly, DANN can provide non-trivial adversarial robustness
under the transfer attack and even the FPA attack (which is better than RMC).
However, it is broken by our GMSA-MIN attack. For example, on MNIST,
the robustness is reduced from $96.66\%$ to $6.17\%$, and on CIFAR-10,
the robustness is reduced from $8.55\%$ to $0.08\%$.

\begin{table*}[t!]
  \small
  \begin{adjustbox}{width=\columnwidth,center}
    \begin{tabular}{l|l|l|c|cccc}
      \toprule
      \multirow{2}{0.08\linewidth}{\textbf{Dataset}} & \multirow{2}{0.12\linewidth}{\textbf{Setting}} & \multirow{2}{0.08\linewidth}{\textbf{Method}} &  \multirow{2}{0.1\linewidth}{\textbf{Accuracy}} & \multicolumn{4}{c}{\textbf{Robustness}}  \\ \cline{5-8} 
                                                     &  &  &  & PGD & FPA & GMSA-AVG & GMSA-MIN \\  \hline \hline 
      \multirow{4}{0.08\linewidth}{{\bf MNIST}} & \multirow{2}{0.12\linewidth}{Inductive} & Standard & 99.42 & 0.00 & - & - & - \\ 
                                                     & & Madry et al. & 99.16 & 91.61 & - & - & - \\ \cline{2-8}			
                                                     & \multirow{2}{0.12\linewidth}{Transductive} & DANN & 99.27 & 96.66 & 96.81 & 79.37 & 6.17 \\ 
                                                     & & ATRM & 99.02 & 95.55 & 95.15 & 94.32 & 95.22 \\ \hline
      \multirow{4}{0.08\linewidth}{{\bf CIFAR-10}} & \multirow{2}{0.12\linewidth}{Inductive} & Standard & 93.95 & 0.00 & - & - & - \\ 
                                                     &  & Madry et al. & 86.06 & 41.06 & - & - & - \\ \cline{2-8}
                                                     & \multirow{2}{0.12\linewidth}{Transductive} & DANN & 92.05 & 54.29 & 8.55 & 0.51 & 0.08 \\  
                                                     &  & ATRM & 85.11 & 60.71 & 61.59 & 53.53 & 57.66 \\  
      \bottomrule
    \end{tabular}
  \end{adjustbox}
  \caption[]{\small Results of DANN and ATRM,
    and comparisons with the baselines in the inductive setting.
    For fair comparison, different learning methods share the same model architecture
    and basic training configuration.
    The PGD attack for DANN and ATRM is the transfer attack.}
  
  \label{tab:dann-atrm-eval-exp}
\end{table*}

\textbf{Effects of combining adversarial training with UDA}.
Table~\ref{tab:dann-atrm-eval-exp} report results for ATRM.
Under our strongest adaptive attacks, ATRM still provides significant adversarial robustness,
and improves over \emph{adversarial training alone}:
On MNIST, ATRM improves the robustness from $91.61\%$ to $94.32\%$;
on CIFAR-10, it improves from $41.06\%$ to $53.53\%$.
These encouraging results suggest further exploration of the utility of transductive
learning for adversarial robustness is warranted.


%% file: appendix.tex
\begin{center}
	\textbf{\LARGE Supplementary Material}
\end{center}

 \begin{center}
	\textbf{\large Towards Adversarial Robustness via Transductive Learning}
\end{center}

We introduce the related work in Section~\ref{sec:related} and the threat model for classic adversarial robustness in Section~\ref{sec:minimax-model}. In Section~\ref{sec:theory-app}, we present our theoretical results on transductive defenses and their proofs. In Section~\ref{sec:exp-details}, we describe the detailed settings for the experiments and also present some additional experimental results.   


\section{Related Work}
\label{sec:related}
\input{related}

\section{Threat Model for Classic Adversarial Robustness}
\label{sec:minimax-model}
\begin{mdframed}[backgroundcolor=black!10, nobreak=true]
  \small
  \begin{definition}[{\bf Threat model for classic adversarial robustness}]
    \label{def:minimax-model}
    Attacker and defender agree on a particular attack type.
    Attacker is an algorithm $\calA$,
    and defender is a supervised learning algorithm $\calT$.\\

    \hrule
    \noindent\centerline{\bf Before the game}\\
    \noindent{\bf Data setup}\\
    $\bullet$\ \ A (labeled) training set $D$ is sampled i.i.d.\ from $(X, Y)$.
    \vskip 3pt
    \hrule

    \noindent\centerline{\bf During the game}\\
    \noindent{\bf Training time}\\
    $\bullet$\ \ (\textbf{Defender}) Train a model $F$ on $D$ as $F = \calT(D)$.

    \noindent{\bf Test time}\\
    $\bullet$\ \ A (labeled) natural test set $V$ is sampled i.i.d.\ from $(X, Y)$.\\
    $\bullet$\ \ (\textbf{Attacker}) On input $F$, $D$, and $V$,
    $\calA$ perturbs each point $(x,y) \in V$ to $(x', y)$
    (subject to the agreed attack type, i.e. $x' \in N(x)$),
    giving $\widetilde{V}=\calA(F, D, V)$.
    \vskip 3pt
    \hrule

    \noindent\centerline{\bf After the game}\\
    \noindent{\bf Evaluation (referee)}\\
    Evaluate the test loss of $F$ on $\widetilde{V}$, $L(F, \widetilde{V})$.
    Attacker's goal is to maximize the test loss,
    while the defender's goal is to minimize the test loss.
  \end{definition}
\end{mdframed}

\section{Theoretical Results on Transductive Defenses}
\label{sec:theory-app}
\input{appendix/theoretical-evidence}

\section{Experimental Details}
\label{sec:exp-details}

\subsection{General Setup}
\label{sec:general-exp-setup}

\subsubsection{Computing Infrastructure}
We run all experiments with PyTorch and NVIDIA GeForce RTX 2080Ti GPUs.

\subsubsection{Dataset}
\label{sec:dataset}
We use three datasets GTSRB, MNIST, and CIFAR-10 in our experiments. The details about these datasets are described below.

\textbf{GTSRB. } The German Traffic Sign Recognition Benchmark (GTSRB)~\cite{stallkamp12} is a dataset of color images depicting 43 different traffic signs. The images are not of a fixed dimensions and have rich background and varying light conditions as would be expected of photographed images of traffic signs. There are about 34,799 training images, 4,410 validation images and 12,630 test images. We resize each image to $32 \times 32$. The dataset has a large imbalance in the number of sample occurrences across classes. We use data augmentation techniques to enlarge the training data and make the number of samples in each class balanced. We construct a class preserving data augmentation pipeline consisting of rotation, translation, and projection transforms and apply this pipeline to images in the training set until each class contained 10,000 examples. We also preprocess images via image brightness normalization and normalize the range of pixel values to $[0, 1]$. 

\textbf{MNIST. } The MNIST~\cite{lecun1998mnist} is a large dataset of handwritten digits. Each digit has 5,500 training images and 1,000 test images. Each image is a $28 \times 28$ grayscale. We normalize the range of pixel values to $[0, 1]$. 

\textbf{CIFAR-10. } The CIFAR-10~\cite{krizhevsky2009learning} is a dataset of 32x32 color images with ten classes, each consisting of 5,000 training images and 1,000 test images. The classes correspond to dogs, frogs, ships, trucks, etc. We normalize the range of pixel values to $[0, 1]$. 

\subsubsection{Implementation of Greedy Model Space Attack (GMSA)}
\label{sec:gmsa-implementation-detail}
We use the Projected Gradient Decent (PGD)~\cite{MMSTV18} to solve the attack objective of GMSA. We apply PGD for each data point in $V$ independently to compute the adversarial perturbation for the data point. For GMSA-AVG, at the $i$-th iteration, when applying PGD on the data point $\bfx$ to generate the perturbation $\delta$, we need to do one backpropagation operation for each model in $\{ F^{(j)} \}_{j=0}^i$ per PGD step. So we do $i+1$ times backpropagation in total. We do the backpropagation for each model sequentially and then accumulate the gradients to update the perturbation $\delta$ since we might not have enough memory to store all the models and compute the gradients at once, especially when $i$ is large. For GMSA-MIN, at the $i$-th iteration, when applying PGD on the data point $\bfx$ to generate the perturbation $\delta$, we only need to do one backpropagation operation for the model $F^{(j^*)}$ with the minimum loss per PGD step. Here, $j^* = \argmin_{0 \leq j \leq i} L(F^{(j)}, \bfx + \delta)$. We scale the number of PGD steps at the $i$-th iteration by a factor of $i+1$ for GMSA-MIN so that it performs the same number of backpropagation operations as GMSA-AVG in each iteration.                

\subsection{Setup for URejectron Experiments}

We use a subset of the GTSRB augmented training data for our experiments, which has 10 classes and contains 10,000 images for each class. We implement URejectron~\cite{GKKM20} on this dataset using the ResNet18 network~\cite{He16} in the transductive setting. Following~\cite{GKKM20}, we implement the basic form of the URejectron algorithm, with $T=1$ iteration. That is we train a discriminator $h$ to distinguish between examples from $P$ and $Q$, and train a classifier $F$ on $P$. Specifically, we randomly split the data into a training set $D_{\text{train}}$ containing 63,000 images, a validation set $D_{\text{val}}$ containing 7,000 images and a test set $D_{\text{test}}$ containing 30,000 images. We then use the training set $D_{\text{train}}$ to train a classifier $F$ using the ResNet18 network. We train the classifier $F$ for 10 epochs using Adam optimizer with a batch size of 128 and a learning rate of $10^{-3}$. The accuracy of the classifier on the training set $D_{\text{train}}$ is 99.90\% and its accuracy on the validation set $D_{\text{val}}$ is 99.63\%. We construct a set $\tilde{x}$ consisting of 50\% normal examples and 50\% adversarial examples. The normal examples in the set $\tilde{x}$ form a set $z$. We train the discriminator $h$ on the set $D_{\text{train}}$ (with label 0) and the set $\tilde{x}$ (with label 1). We then evaluate URejectron's performance on $\tilde{x}$: under a certain threshold used by the discriminator $h$, we measure the fraction of normal examples in $z$ that are rejected by the discriminator $h$ and the error rate of the classifier $F$ on the examples in the set $\tilde{x}$ that are accepted by the discriminator $h$. The set $z$ can be $D_{\text{test}}$ or a set of corrupted images generated on $D_{\text{test}}$. We use the method proposed in~\cite{Hendrycks19} to generate corrupted images with the corruption type of brightness and the severity level of 1. The accuracy of the classifier on the corrupted images is 98.90\%. The adversarial examples in $\tilde{x}$ are generated by the PGD attack~\cite{MMSTV18} or the CW attack~\cite{CW17}. For PGD attack, we use $L_\infty$ norm with perturbation budget $\epsilon=8/255$ and random initialization. The number of iterations is 40 and the step size is $1/255$. The robustness of the classifier under the PGD attack is 3.66\%. For CW attack, we use $L_2$ norm as distance measure and set $c=1$ and $\kappa=0$. The learning rate is 0.01 and the number of steps is 100. The robustness of the classifier under the CW attack is 0.00\%.   

\subsection{Setup for RMC Experiments}

\begin{table*}[t!]
        \centering
		\begin{tabular}{l|l|c|c}
			\toprule
			 \textbf{Dataset} & \textbf{Model} & \textbf{Accuracy} &  \textbf{Robustness}  \\ \hline \hline 
			\multirow{2}{0.08\linewidth}{{{\bf MNIST}}}  
            & Standard  & 99.50 & 0.00 \\  
			& Madry et al. & 99.60 & 93.50 \\ \hline
			\multirow{2}{0.08\linewidth}{{{\bf CIFAR-10}}}  
            & Standard  & 94.30 & 0.00 \\  
			& Madry et al. & 83.20 & 46.80 \\  
			\bottomrule
		\end{tabular}
	\caption[]{\small Performance of the pre-trained models used by RMC. The robustness of the models is evaluated under the PGD attack. All values are percentages. }
	\label{tab:rmc-base-model-performance}
\end{table*}

We follow the settings in~\cite{Wu20} and perform experiments on MNIST and CIFAR-10 datasets to evaluate the adversarial robustness of RMC. Under our evaluation framework, RMC can be treated as an adaptation method $\Gamma$ and the size of the test set $U$ is 1 since RMC adapts the model based on a single data point. Suppose $V=\{(\hat{\bfx}, y) \}$, $U=\{\hat{\bfx}\}$ and the current model is $F$, then the loss of RMC on $U$ is $L(\Gamma(U, F), V)$. Given a sequence of test inputs $(\hat{\bfx}^{(1)}, y^{(1)}),\cdots, (\hat{\bfx}^{(n)}, y^{(n)})$, suppose the initial model is $F^{(0)}$, then the loss of RMC on the data sequence is $\frac{1}{n} \sum_{i=1}^{n} L(F^{(i)}, \hat{\bfx}^{(i)}, y^{(i)})$, where $F^{(i)} = \Gamma(F^{(i-1)}, \hat{\bfx}^{(i)})$. We assume that the attacker knows $\Gamma$ and can use it to simulate the adaptation process to generate a sequence of adversarial examples $\hat{\bfx}^{(1)},\cdots, \hat{\bfx}^{(n)}$. Then we evaluate the robustness of RMC on the generated data sequence $\hat{\bfx}^{(1)},\cdots, \hat{\bfx}^{(n)}$. 

We also evaluate RMC and RMC$^+$ (an extended version of RMC) under the PGD-skip attack setting proposed in~\cite{Wu20}: the attacker generates an adversarial example $\hat{\bfx}^{(p+1)}$ against the network that has been adapted to $\hat{\bfx}^{(1)},\cdots, \hat{\bfx}^{(p)}$. We follow their original setting for PGD-skip: first generate adversarial examples $\hat{\bfx}^{(1)},\cdots, \hat{\bfx}^{(p)}$ using the initial model $F^{(0)}$, and then generate the adversarial example $\hat{\bfx}^{(p+1)}$ on a clean input $\bfx^{(p+1)}$ randomly sampled from the test data distribution using the model $F^{(p)}$ that has been adapted to $\hat{\bfx}^{(1)},\cdots, \hat{\bfx}^{(p)}$. The robustness of RMC (or RMC$^+$) is evaluated on $\hat{\bfx}^{(p+1)}$ using $F^{(p)}$. We repeat the experiment independently for 1000 times and calculate the average robustness. To save computational cost, we use the same $\hat{\bfx}^{(1)},\cdots, \hat{\bfx}^{(p)}$ for all independent experiments. 

We consider two kinds of pre-trained models for RMC (or RMC$^+$): one is the model trained via standard supervised training; the other is the model trained using the adversarial training proposed in~\cite{MMSTV18}. The performance of the pre-trained models is shown in Table~\ref{tab:rmc-base-model-performance}. We describe the settings for each dataset below. 

\subsubsection{MNIST} 

\textbf{Model architecture and training configuration. } We use a neural network with two convolutional layers, two full connected layers and batch normalization layers. For both standard training and adversarial training, we train the model for 100 epochs using the Adam optimizer with a batch size of 128 and a learning rate of $10^{-3}$. We use the $L_\infty$ norm PGD attack as the adversary for adversarial training with a perturbation budget $\epsilon$ of $0.3$, a step size of $0.01$, and number of steps of $40$.   

\textbf{RMC and RMC$^+$ configuration. } We set $K=1024$. Suppose the clean training set is $\mathbb{D}$. Let $\mathbb{D}'$ contain $|\mathbb{D}|$ clean inputs and $|\mathbb{D}|$ adversarial examples. So $N'=2|\mathbb{D}|$. We generate the adversarial examples using the $L_\infty$ norm PGD attack with a perturbation budget $\epsilon$ of $0.3$, a step size of $0.01$, and number of steps of $100$. We extract the features from the penultimate layer of the model and use the Euclidean distance in the feature space of the model to find the top-K nearest neighbors of the inputs. When adapting the model, we use Adam as the optimizer and set the learning rate to be $2\times 10^{-4}$. We train the model until the early-stop condition holds. That is the training epoch reaches 100 or the validation loss doesn't decrease for 5 epochs. For RMC$^+$, we use the same configuration, except that we update $\mathbb{D}'$ using the model $F^{(p)}$ when evaluating it on $\bfx^{(p+1)}$.  

\textbf{Attack configuration. } We use PGD to solve the attack objectives of all attacks used for our evaluation, including FPA, GMSA-AVG, GMSA-MIN and PGD-skip. We use the same configuration for all attacks: $L_\infty$ norm PGD with a perturbation budget $\epsilon$ of $0.3$, a step size of $0.01$, and number of steps of $100$. We set $T=9$ for FPA, GMSA-AVG and GMSA-MIN.   

\subsubsection{CIFAR-10} 

\textbf{Model architecture and training configuration. } We use the ResNet-32 network~\cite{He16}. For both standard training and adversarial training, we train the model for 100 epochs using Stochastic Gradient Decent (SGD) optimizer with Nesterov momentum and learning rate schedule. We set momentum $0.9$ and $\ell_2$ weight decay with a coefficient of $10^{-4}$. The initial learning rate is $0.1$ and it decreases by $0.1$ at 50, 75 and 90 epoch respectively. The batch size is $128$. We augment the training images using random crop and random horizontal flip. We use the $L_\infty$ norm PGD attack as the adversary for adversarial training with a perturbation budget $\epsilon$ of $8/255$, a step size of $2/255$, and number of steps of $10$. 

\textbf{RMC and RMC$^+$ configuration. } We set $K=1024$. Suppose the clean training set is $\mathbb{D}$. Let $\mathbb{D}'$ contain $|\mathbb{D}|$ clean inputs and $4|\mathbb{D}|$ adversarial examples. So $N'=5|\mathbb{D}|$. We generate the adversarial examples using the $L_\infty$ norm PGD attack with a perturbation budget $\epsilon$ of $8/255$, a step size of $1/255$, and number of steps of $40$. We extract the features from the penultimate layer of the model and use the Euclidean distance in the feature space of the model to find the top-K nearest neighbors of the inputs. We use Adam as the optimizer and set the learning rate to be $2.5 \times 10^{-5}$. For RMC$^+$, we use the same configuration, except that we update $\mathbb{D}'$ using the model $F^{(p)}$ when evaluating it on $\bfx^{(p+1)}$.  

\textbf{Attack configuration. } We use PGD to solve the attack objectives of all attacks used for our evaluation, including FPA, GMSA-AVG, GMSA-MIN and PGD-skip. We use the same configuration for all attacks: $L_\infty$ norm PGD with a perturbation budget $\epsilon$ of $8/255$, a step size of $1/255$, and number of steps of $40$. We set $T=9$ for FPA, GMSA-AVG and GMSA-MIN. 

\subsection{Setup for DANN and ATRM Experiments}

We perform experiments on MNIST and CIFAR-10 datasets. We describe the settings for each dataset below. 

\subsubsection{MNIST}

\textbf{Model architecture. } We use the same model architecture as the one used in~\cite{chuang20}, which is shown below. 

\begin{center}
\small
 \begin{tabular}{||c||} 
 \hline
 Encoder  \\ [0.5ex] 
 \hline\hline
 nn.Conv2d(3, 64, kernel$\_$size=5)  \\ 
 \hline
 nn.BatchNorm2d \\
 \hline
 nn.MaxPool2d(2)  \\
 \hline 
 nn.ReLU \\
 \hline
 nn.Conv2d(64, 128, kernel$\_$size=5)  \\ 
 \hline
 nn.BatchNorm2d \\
 \hline
 nn.Dropout2d \\
 \hline
 nn.MaxPool2d(2)  \\
 \hline 
 nn.ReLU \\
 \hline \hline
 nn.Conv2d(128, 128, kernel$\_$size=3, padding=1)  \\
 \hline
 nn.BatchNorm2d \\
 \hline
 nn.ReLU \\
 \hline
 $\times 2$ \\
 \hline
\end{tabular}
\quad

\begin{tabular}{||c||} 
 \hline
 Predictor  \\ [0.5ex] 
 \hline\hline
 nn.Conv2d(128, 128, kernel$\_$size=3, padding=1)  \\
 \hline
 nn.BatchNorm2d \\
 \hline
 nn.ReLU \\
 \hline
 $\times 3$ \\
 \hline\hline
 flatten \\
 \hline
 nn.Linear(2048, 256)  \\ 
 \hline 
 nn.BatchNorm1d \\
 \hline
 nn.ReLU  \\
 \hline
 nn.Linear(256, 10) \\
 \hline 
 nn.Softmax \\
 \hline
\end{tabular}
\quad
 \begin{tabular}{||c||} 
 \hline
 Discriminator  \\ [0.5ex] 

 \hline\hline
 nn.Conv2d(128, 128, kernel$\_$size=3, padding=1)  \\
 \hline
 nn.ReLU \\
 \hline
 $\times 5$ \\
 \hline\hline
 Flatten \\
 \hline
 nn.Linear(2048, 256)  \\ 
 \hline 
 nn.ReLU  \\
 \hline 
 nn.Linear(256, 2) \\
 \hline 
 nn.Softmax \\
 \hline
\end{tabular}

\end{center}

\textbf{Training configuration. } We train the models for 100 epochs using the Adam optimizer with a batch size of 128 and a learning rate of $10^{-3}$. We use the $L_\infty$ norm PGD attack as the adversary to generate adversarial training examples with a perturbation budget $\epsilon$ of $0.3$, a step size of $0.01$, and number of steps of $40$. For the representation matching in DANN and ATRM, we adopt the original progressive training strategy for the discriminator~\cite{ganin16} where the weight $\alpha$ for the domain-invariant loss is initiated at 0 and is gradually changed to 0.1 using the schedule $\alpha=\frac{2}{1+\text{exp}(-10\cdot p)}-1$, where $p$ is the training progress linearly changing from 0 to 1.

\textbf{Attack configuration. } We use PGD to solve the attack objectives of all attacks used for our evaluation, including the transfer attack, FPA, GMSA-AVG, and GMSA-MIN. We use the same configuration for all attacks: $L_\infty$ norm PGD with a perturbation budget $\epsilon$ of $0.3$, a step size of $0.01$, and number of steps of $200$. We set $T=9$ for FPA, GMSA-AVG and GMSA-MIN. When attacking DANN, we use the model trained via standard training as the initial model $F^{(0)}$ for the transfer attack, FPA and GMSA; when attacking ATRM, we use the model trained with adversarial training as the initial model $F^{(0)}$ for the transfer attack, FPA and GMSA. 

\subsubsection{CIFAR-10}

\textbf{Model architecture. } We use the ResNet-18 network~\cite{He16} and extract the features from the third basic block for representation matching. The detailed model architecture is shown below. 

\begin{center}
\small
 \begin{tabular}{||c||} 
 \hline
 Encoder  \\ [0.5ex] 
 \hline\hline
 nn.Conv2d(3, 64, kernel$\_$size=3)  \\ 
 \hline
 nn.BatchNorm2d \\
 \hline 
 nn.ReLU \\
 \hline
 BasicBlock(in$\_$planes=64, planes=2, stride=1) \\
 \hline 
 BasicBlock(in$\_$planes=128, planes=2, stride=2) \\
 \hline 
 BasicBlock(in$\_$planes=256, planes=2, stride=2) \\
 \hline
\end{tabular}
\quad

\begin{tabular}{||c||} 
 \hline
 Predictor  \\ [0.5ex] 
 \hline\hline
 BasicBlock(in$\_$planes=512, planes=2, stride=2)  \\
 \hline
 avg$\_$pool2d \\
 \hline
 flatten \\
 \hline
 nn.Linear(512, 10)  \\ 
 \hline
 nn.Softmax \\
 \hline
\end{tabular}
\quad
 \begin{tabular}{||c||} 
 \hline
 Discriminator  \\ [0.5ex] 
\hline\hline
 BasicBlock(in$\_$planes=512, planes=2, stride=2)  \\
 \hline
 avg$\_$pool2d \\
 \hline
 flatten \\ \hline
 nn.Linear(512, 2)  \\ 
 \hline
 nn.Softmax \\
 \hline
\end{tabular}

\end{center}

\textbf{Training configuration. } We train the models for 100 epochs using stochastic gradient decent (SGD) optimizer with Nesterov momentum and learning rate schedule. We set momentum $0.9$ and $\ell_2$ weight decay with a coefficient of $10^{-4}$. The initial learning rate is $0.1$ and it decreases by $0.1$ at 50, 75 and 90 epoch respectively. The batch size is $64$. We augment the training images using random crop and random horizontal flip. We use the $L_\infty$ norm PGD attack as the adversary to generate adversarial training examples with a perturbation budget $\epsilon$ of $8/255$, a step size of $2/255$, and number of steps of $5$. For the representation matching in DANN and ATRM, we adopt the original progressive training strategy for the discriminator~\cite{ganin16} where the weight $\alpha$ for the domain-invariant loss is initiated at 0 and is gradually changed to 0.1 using the schedule $\alpha=\frac{2}{1+\text{exp}(-10\cdot p)}-1$, where $p$ is the training progress linearly changing from 0 to 1.

\textbf{Attack configuration. } We use PGD to solve the attack objectives of all attacks used for our evaluation, including the transfer attack, FPA, GMSA-AVG, and GMSA-MIN. We use the same configuration for all attacks: $L_\infty$ norm PGD with a perturbation budget $\epsilon$ of $8/255$, a step size of $1/255$, and number of steps of $100$. We set $T=9$ for FPA, GMSA-AVG and GMSA-MIN. When attacking DANN, we use the model trained via standard training as the initial model $F^{(0)}$ for the transfer attack, FPA and GMSA; when attacking ATRM, we use the model trained with adversarial training as the initial model $F^{(0)}$ for the transfer attack, FPA and GMSA. 

\subsection{Detailed Results for Attacking DANN and ATRM}
\label{sec:dann-detail-attack-results}

\begin{figure*}[t]
  \centering
  \begin{subfigure}{0.4\linewidth}
    \centering
    \includegraphics[bb=0 0 282 228,width=\linewidth]{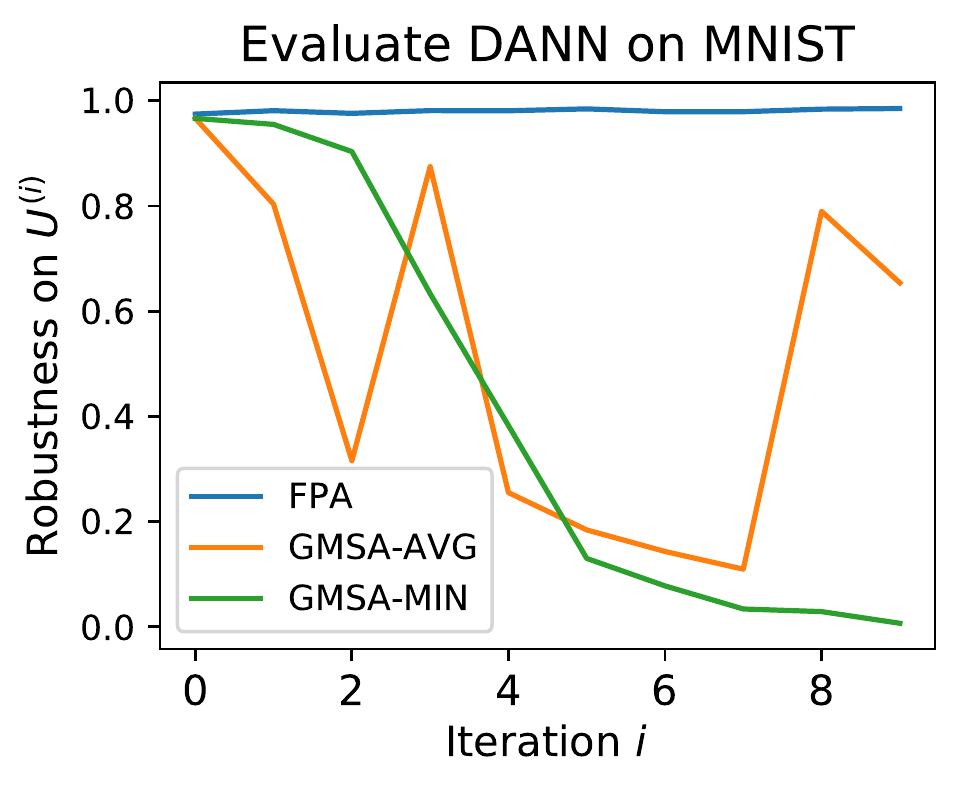}
  \end{subfigure}
  \begin{subfigure}{0.4\linewidth}
    \centering
    \includegraphics[bb=0 0 282 228,width=\linewidth]{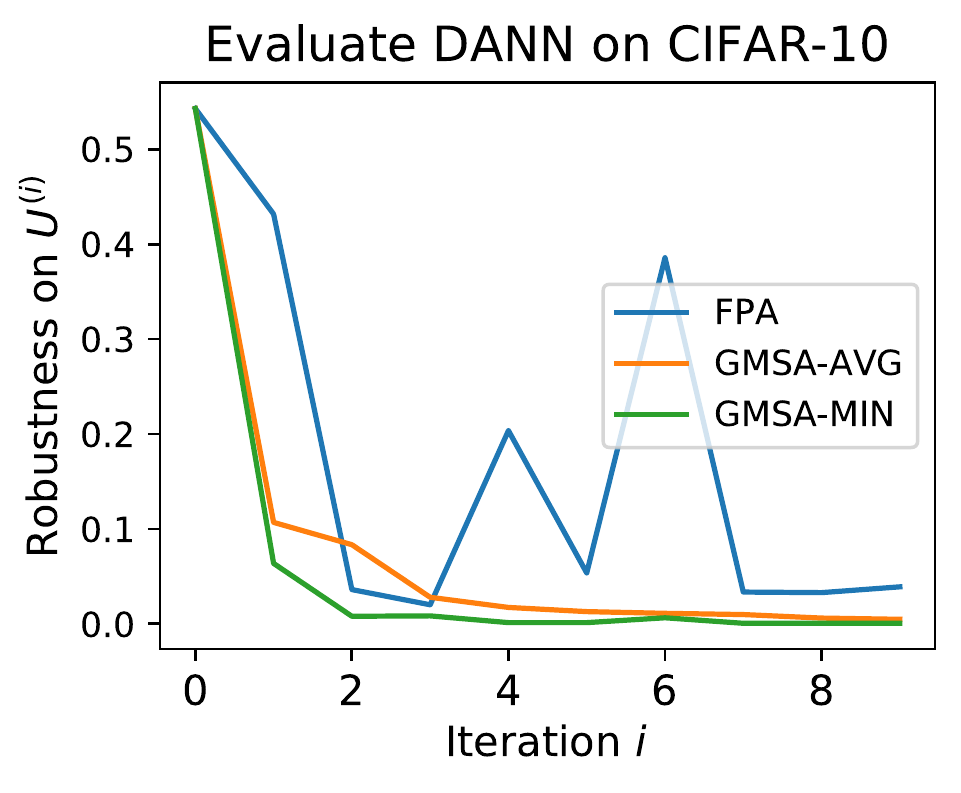}
  \end{subfigure}
 \begin{subfigure}{0.4\linewidth}
    \centering
    \includegraphics[bb=0 0 282 228,width=\linewidth]{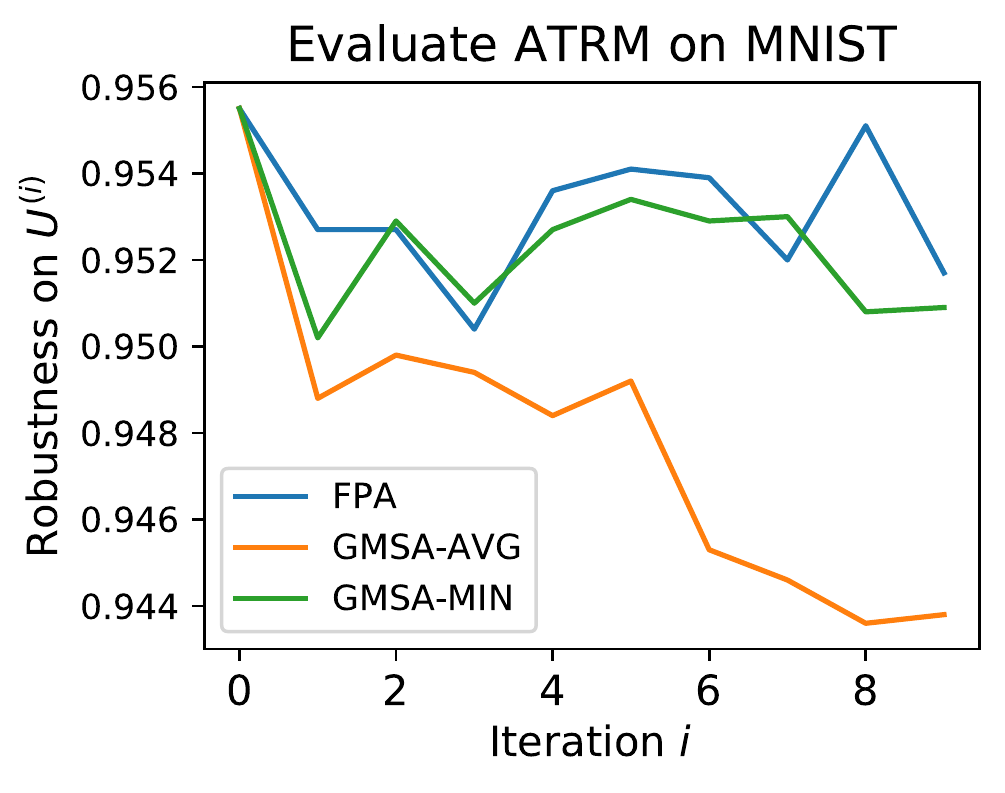}
  \end{subfigure} 
  \begin{subfigure}{0.4\linewidth}
    \centering
    \includegraphics[bb=0 0 282 228,width=\linewidth]{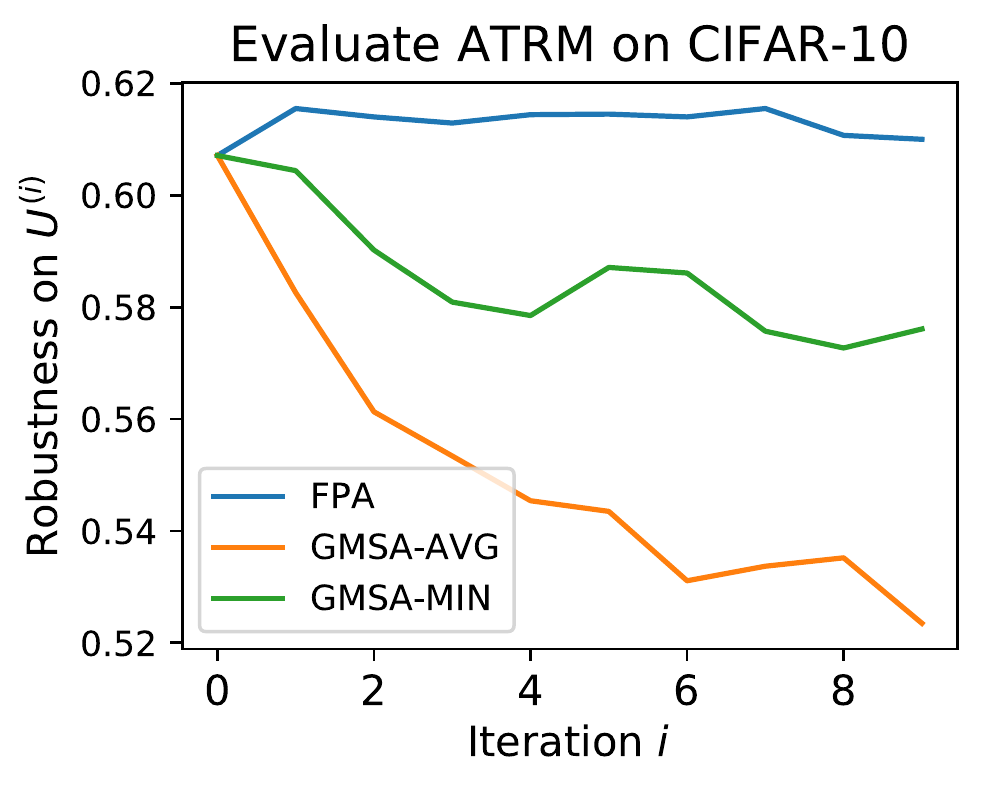}
  \end{subfigure}
  \caption{\small Detailed results for attacking DANN and ATRM on MNIST and CIFAR-10 using FPA, GMSA-AVG and GMSA-MIN attacks. The robustness of DANN (or ATRM) is evaluated by the attacker with his own randomness.      
  }
  \label{fig:dann-atrm-attack-detail-exp}
\end{figure*}

Figure~\ref{fig:dann-atrm-attack-detail-exp} shows the robustness of DANN and ATRM on the perturbed set $U^{(i)}$ generated by our attacks (FPA, GMSA-AVG, and GMSA-MIN) for each iteration $i$. Note that the robustness here is computed by the attacker with his randomness, which is different from the defender's private randomness. The results show that usually FPA is not effective in attacking DANN and ATRM, and it cannot generate increasingly stronger attack sets over iterations. In attacking DANN, our GMSA-MIN attack is more effective and can generate increasing stronger attack sets over iterations while in attacking ATRM, our GMSA-AVG attack is more effective and can also generate increasing stronger attack sets over iterations. Compared to the robustness achieved by the defender shown in Table~\ref{tab:dann-atrm-eval-exp}, we can see that by using private randomness, the defender may be able to achieve better robustness. For example, on MNIST, the robustness of DANN on the strongest attack set generated by GMSA-AVG is only 10.97\%, but the defender can achieve 79.37\% robustness on this attack set with his private randomness. This is because the attack set fails to attack the model space of the DANN defense (we observe that on this attack set, some previous models $F^{(i)}$ can achieve 71.28\% robustness).     

\subsection{The Effect of Different Private Randomness for DANN and ATRM}
\label{sec:effect-of-randomness}
\begin{figure*}[t]
  \centering
  \begin{subfigure}{0.4\linewidth}
    \centering
    \includegraphics[bb=0 0 286 209,width=\linewidth]{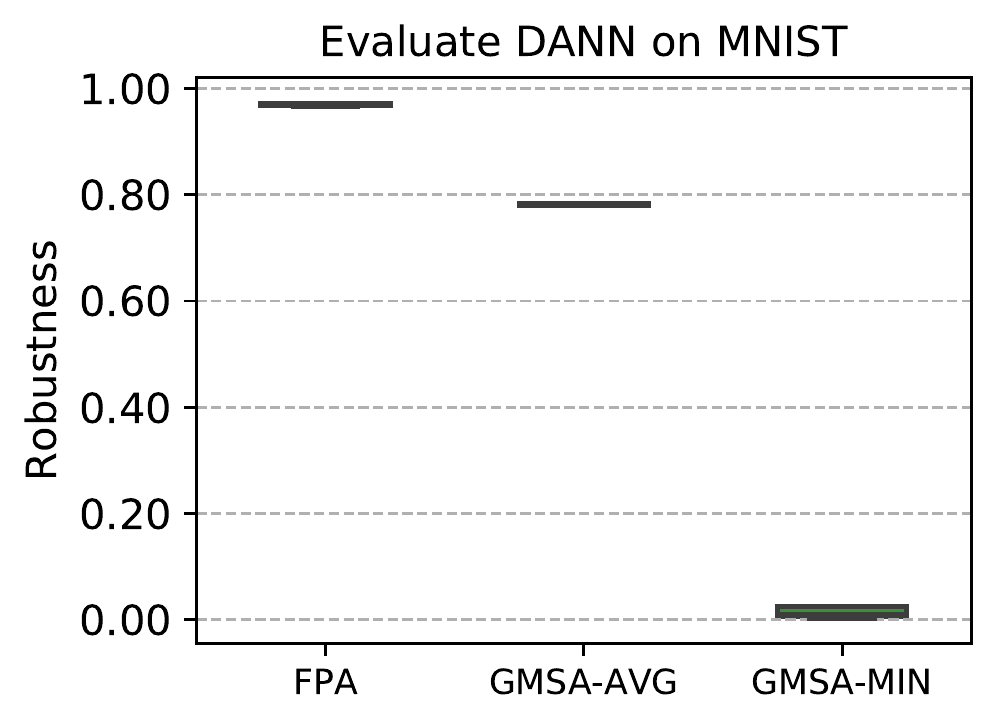}
  \end{subfigure}
  \begin{subfigure}{0.4\linewidth}
    \centering
    \includegraphics[bb=0 0 286 209,width=\linewidth]{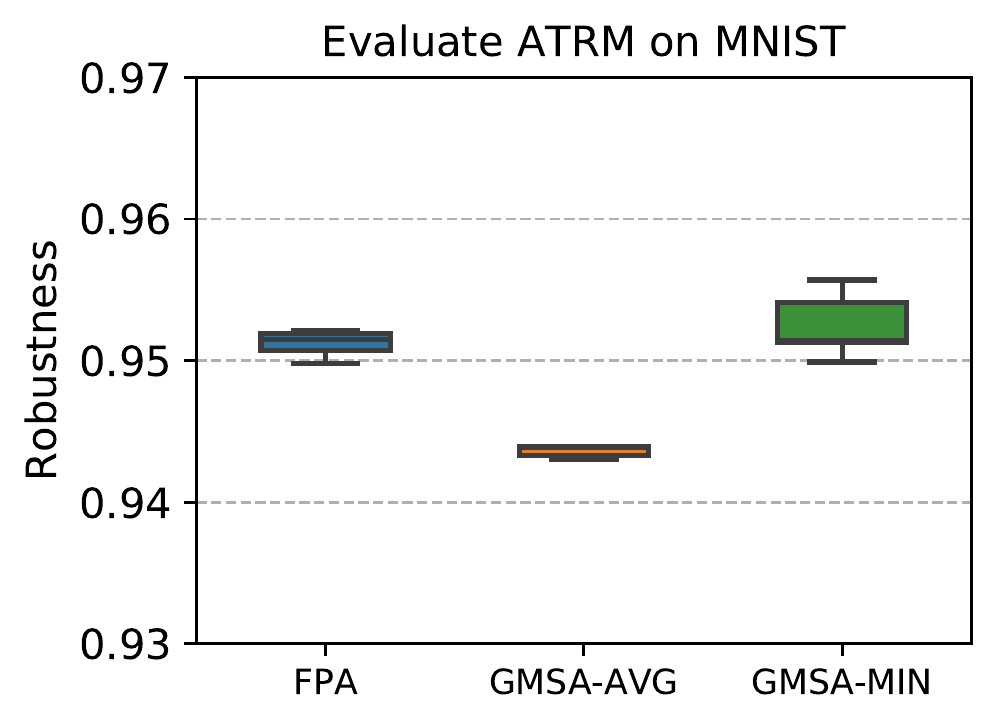}
  \end{subfigure}
 \begin{subfigure}{0.4\linewidth}
    \centering
    \includegraphics[bb=0 0 286 209,width=\linewidth]{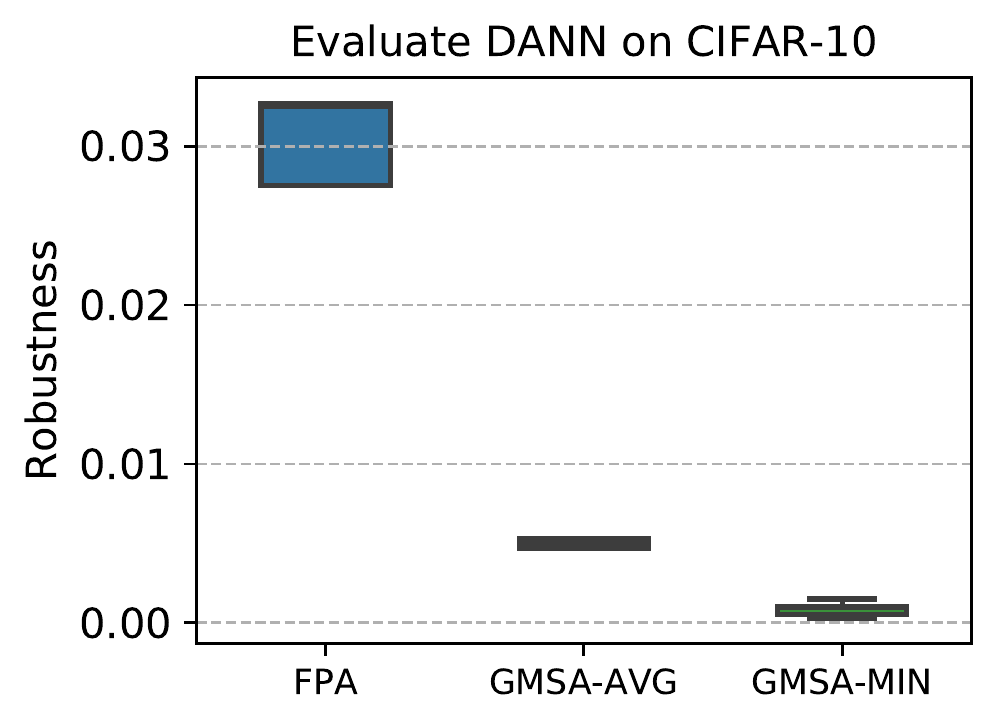}
  \end{subfigure} 
  \begin{subfigure}{0.4\linewidth}
    \centering
    \includegraphics[bb=0 0 286 209,width=\linewidth]{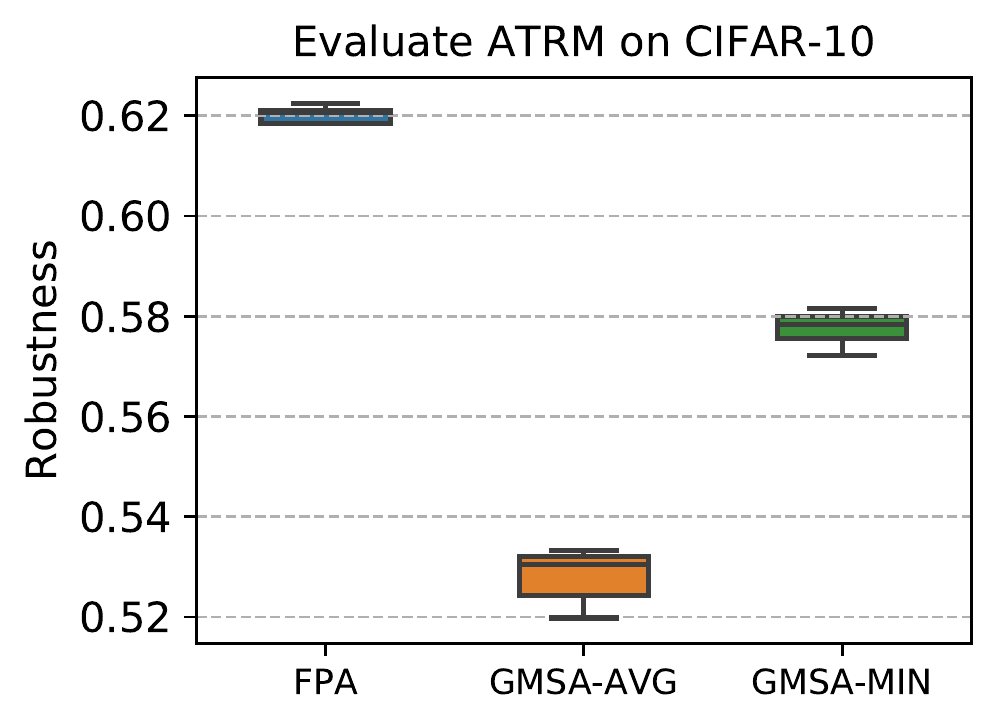}
  \end{subfigure}
  \caption{\small Multiple runs of DANN and ATRM defense experiments with different random seeds. 
  }
  \label{fig:multiple-runs-defense-exp}
\end{figure*}

We run the DANN and ATRM defense experiments five times with different random seeds on the same strongest attack set generated by our attacks (FPA, GMSA-AVG, and GMSA-MIN). The results in Figure~\ref{fig:multiple-runs-defense-exp} show that the robustness of DANN and ATRM doesn't vary much with different private randomness. 

%% file: related.tex
This paper presents an interplay between three research directions:
Adversarial robustness, transductive learning, and domain adaptation.

\textbf{\em Adversarial robustness in the inductive setting.}
Many attacks have been proposed to evaluate the adversarial robustness of the defenses
in the inductive setting where the model is fixed during the evaluation phase~\cite{Goodfellow15,CW17,Kurakin17,Moosavi16,Croce20}.
Principles for adaptive attacks have been developed in~\cite{Tram20} and many existing defenses are shown to be broken
based on attacks developed from these principles~\cite{ACW18}.
A fundamental method to obtain adversarial robustness in this setting is
adversarial training~\cite{MMSTV18,Zhang19}.

\textbf{\em Adversarial robustness in the transductive setting.}
There have been emerging interests in researching dynamic model defenses in the transductive setting.
A research agenda for dynamic model defense has been proposed~\cite{G19}.
\cite{Wu20} proposed the first defense method called RMC under this setting to
improve the adversarial robustness of a model after deployment.
However, their attacks for evaluation are somewhat weak
according to our principle of attacking model space for transductive defense and in this work,
we show that one can indeed break their transductive defense by attacking model space. 

\textbf{\em Domain adaptation methods.}
Domain adaptation is a set of techniques for training models
where the target domain differs from the source. DANN~\cite{ajakan2014domain} is a classic technique
for unsupervised domain adaptation (UDA) where we have access to unlabeled test data.
In this work, we propose a novel use of DANN as a transductive defense method.
We showed that DANN alone is susceptible to model space attacks,
but gives a nontrivial improvement of adversarial robustness when combined with adversarial training. 

\textbf{\em Test-time adaptation methods.}
Test-time adaptation is a recent paradigm to further improve the efficiency
for adapting to novel domains~\cite{wang2021tent,SWZMHE20}.
However, these methods are not designed to achieve adversarial robustness
and we find that they are vulnerable even under the transfer attacks.


%% file: appendix/theoretical-evidence.tex
\subsection{Valuation of the Game}
\label{sec:valuation}
\begin{proposition}[{\bf maximin vs. classic minimax threat model}]
  \label{prop:no-harder}
  Let $k \ge 1$ be a natural number, and $\calF$ be the hypothesis class.
  For a given $V$, the domain of $\widetilde{V}$
  is a well-defined function of $V$ (e.g., $\ell_\infty$ ball around $V$).
  We have that:
  $\Exp_{V \sim (X, Y)^k}\left[\max_{U}\min_{\widetilde{F} \in \calF}
    \{L(\widetilde{F}, \widetilde{V})\}\right]
  \le \min_{\widetilde{F} \in \calF}\Exp_{V \sim (X, Y)^k}\left[
    \max_{\widetilde{V}} \{L(\widetilde{F}, \widetilde{V})\}\right]$
\end{proposition}
\begin{proof}
  Let $\calF$ be the family of models $\widetilde{F}$ we can choose from.
  From the maximin inequality, we have that
  \begin{align*}
    \max_{U}\min_{\widetilde{F} \in \calF}
    \{L(\widetilde{F}, \widetilde{V})\}
    \le & \min_{\widetilde{F} \in \calF}\max_{\widetilde{V}}
          \{L(\widetilde{F}, \widetilde{V})\}\\
  \end{align*}
  Note that for the minimax, the max over $\widetilde{V}$ is also constrained
  to perturb features (as we want to find adversarial examples).
  If we take expectation over $V$, we have then
  \begin{align*}
    \Exp_V\left[\max_{U}\min_{\widetilde{F} \in \calF}
    \{L(\widetilde{F}, \widetilde{V})\}\right]
    \le \Exp_V\left[\min_{\widetilde{F}\in\calF}\max_{\widetilde{V}}
    \{L(\widetilde{F}, \widetilde{V})\}\right]
  \end{align*}
  Note that
  \begin{align*}
    \Exp_V\left[\min_{\widetilde{F}\in\calF}\max_{\widetilde{V}}
    \{L(\widetilde{F}, \widetilde{V})\}\right]
    \le \min_{\widetilde{F}\in\calF}\Exp_V\left[\max_{\widetilde{V}}
    \{L(\widetilde{F}, \widetilde{V})\}\right],
  \end{align*}
  which completes the proof.\qed
\end{proof}
The proof 
holds verbatim to the more general semi-supervised threat model.
We also note that, in fact, if the concept class has unbounded VC dimension,
then good models always {\em exist} that can fit both $D$ and $V$ perfectly.
So the valuation of the maximin game is actually always $0$:

\begin{proposition}[\textbf{Good models exist with large capacity}]
  \label{prop:unbounded-vc-existence}
  Consider binary classification tasks and that the hypothesis class $\calF$ has infinite
  VC dimension. Then the valuation of the maximin game
\begin{align}
  \Exp_{V \sim (X, Y)^k}\left[\max_{U}\min_{\widetilde{F} \in \calF}
    \{L(\widetilde{F}, \widetilde{V})\}\right]  
\end{align}
  is $0$. That is, perfect models
  always exist to fit $U$.
\end{proposition}

This thus gives a first evidence that that transductive advesarial learning
is strictly easier. We remark that transductive learning here is essential
(differnet models are allowed for different $U$).
We conclude this section by noting the following:
\begin{proposition}[{\bf Good minimax solution is also a good maximin solution}]
  \label{prop:good-minimax-soln-is-also-good-in-maximin}
  Suppose $\calT^*$ is a supervised learning algorithm
  which trains a model $F^*=\calT^*(D)$, where its adversarial gain in the
  adversarial semi-supervised minimax model is bounded by $\kappa$
  (i.e. $\Exp_{V'}[\max_{\widetilde{V'}}L(F^*, \widetilde{V})] \le \kappa.$)
  Then in the maximin threat model, the adversarial gain of the strategy
  $(\calT^*, \Gamma^*)$, where $\Gamma^*(F^*, D, U) = F^* = \calT^*(D)$,
  is also upper bounded by $\kappa$.
\end{proposition}

\subsection{Usefulness of transductive learning}
\label{sec:gaussian}
\input{appendix/gaussian}


%% file: appendix/gaussian.tex
\newcommand{\nins}{n_0}
\newcommand{\rerr}[1]{\mathrm{err}^{\infty,\epsilon}_\textrm{robust}(#1)}
\newcommand{\thres}{\|\mu\|_2 - c \sigma}
\newcommand{\errt}[1]{\mathrm{err}_{#1}}

Having defined the transductive adversarial threat model, a natural next question
is thus to examine the relationship between our threat model and the classic
inductive threat model. A standard way to study this question is via the valuation
of the respective games, where in the transductive threat model it is a maximin game,
and in the inductive model it is a minimax game. To this end, by standard arguments,
we get immediate results such as the transductive model is no harder than
the inductive threat model. We collect these results in Appendix~\ref{sec:valuation}.

We note, however, that valuation of the game does {\em not} give any insight for the
existence of good {\em transductive} defense algorithms, which can only leverage unlabeled data.
In this section we provide a problem instance (i.e., data distributions and number of data points),
and prove that that transductive threat model is {\em strictly easier}
than the inductive threat model for the problem:
{\em In the inductive model no algorithm can achieve a nontrivial error,
  while in the transductive model there are algorithms achieving small errors.}
Since the transductive model is no harder than the inductive model for all problem instances,
and there is a problem instance where the former is strictly easier,
we thus formally establish a separation between the two threat models.
Furthermore, the problem instance we considered is on Gaussian data.
The fact that transductive model is already strictly easier than inductive in this simple problem
provides positive support for potentially the same phenomenon on more complicated data.

\vskip 5pt
\noindent\textbf{Data distributions and the learning task.}
We consider the homogeneous case (the source and target are the same distribution)
and the $\ell_\infty$ attack.
We consider the classic Gaussian data model recently used for analyzing adversarial robustness in~\cite{schmidt2018adversarially,carmon2019unlabeled}:
A binary classification task where $\calX =\mathbb{R}^d$
and $\calY = \{+1, -1\}$, $y$ uniform on $\mathcal{Y}$ and
$x | y \sim \mathcal{N}(y\mu, \sigma^2 I)$ for a vector $\mu \in \mathbb{R}^d$ with $
\|\mu\|^2_2 = d$
and coordinate noise variance $\sigma^2 > 0$. In words, this is a mixture of two
Gaussians, one with label $+1$, and one with label $-1$.
For both threat models, the datasets $D = \{(x_i,y_i)\}_{i=1}^n$ and $V = \{(x,y)\}$.
In particular, $V$ only has one data point. In the transductive threat model,
we let $x'$ denote the perturbed input obtained from ${x}$ by the $l_\infty$ attack
with bounded norm $\epsilon > 0$, i.e., $x' = {x} + \nu$ with $\|\nu\|_\infty \le \epsilon$.
Put $\widetilde{V} = \{(x',y)\}$ and $U=\{x'\}$. We prove the following:


\begin{theorem}[\textbf{Separation of transductive and inductive threat models}]
  \label{thm:gaussian}
  There exists absolute constants $C_1, C_2, c, C>0$ such that
  for any $\nu \in (0, 1)$, if $\sigma^2 = C_1\sqrt{d \log\frac{1}{\nu}}$, $\frac{d}{\log^2 d} \ge C_2 \left(\frac{1}{\nu} + \frac{1}{\epsilon^4} \log^3\frac{1}{\nu}\right)$, and $c\log^2\frac{1}{\nu} \le n \le \frac{C\epsilon^2}{\log d} \sqrt{d \log \frac{1}{\nu}} $ in the above data model, then:%
  \footnote{The bound on $d$ makes sure the range of $n$ is not empty. The expectation of the error is over the randomness of
    $D, V, V'$, and possible algorithm randomness.}
  \begin{itemize}
  \item[(1)] In the inductive threat model,
    the learned model $\widetilde{F} = \Gamma(\mathcal{T}(D), D, U)$
    by {\em any} algorithms $\mathcal{T}$ and $\Gamma$ must have a large error:
    $\mathbb{E} \left\{ L(\widetilde{F}, \widetilde{V'}) \right\}
    \ge \frac{1}{2} (1- \nu)$.
  \item[(2)] In the transductive threat model, there exist
    $\mathcal{T}$ and $\Gamma$ such that 
    the adapted model $\widetilde{F} = \Gamma(\mathcal{T}(D), D, U)$ has a small error:
    $\mathbb{E} \left\{ L(\widetilde{F}, \widetilde{V}) \right\} \le \nu.$ 
  \end{itemize}
\end{theorem}

In the inductive model, the algorithm needs to estimate $\mu$ to a small error which is not possible with limited data $n \le \frac{C\epsilon^2}{\log d} \sqrt{d \log \frac{1}{\nu}}$ (formally proved via a reduction to the lower bound in~\cite{schmidt2018adversarially}). 
In the transductive model, the algorithm does not need to learn a function that works well for the whole distribution, but only need to search for one that works on $x'$. This allows to search in a much smaller set of hypotheses and requires less training data. In particular, we first use $\Theta(\log^2 \frac{1}{\nu})$ data in $D$ to train a linear classifier $\mathrm{sign}(\hat{\theta}^\top x)$ with a parameter $\hat{\theta}$. Upon receiving $x'$, we construct two large-margin classifiers: in the span of $\hat\theta$ and $x'$, find $\bar\theta_+$ and $\bar\theta_-$ that classify $x'$ as $+1$ and $-1$ with a chosen margin, respectively. Finally, we use another set of $\Theta(\log^2 \frac{1}{\nu})$ data from $D$ to check the two classifiers and pick the one with smaller errors The picked classifier will classify $x'$ correctly w.h.p., though it will not have a small error on the whole data distribution. 
Intuitively, the transductive model allows the algorithm to adapt to the given $U$ and only search for hypotheses that can classifier $U$ correctly. Such adaptivity thus separates the two threat models.

\subsection{Proof of Theorem~\ref{thm:gaussian}}
\label{sec:gaussian-proof}
\input{appendix/gaussian-proof}


%% file: appendix/gaussian-proof.tex
We prove this theorem by a series of lemmas.  Let $\nins := \sigma^4/d $, and choose $K$ such $K \ge \Omega\left(\log\frac{1}{\nu}\right)$ and $K \le \nins$, e.g., $K = \nins$. Note that if $c\log^2\frac{1}{\nu} \le n \le \frac{C\epsilon^2}{\log d} \sqrt{d \log \frac{1}{\nu}} $ for sufficiently large $c$ and $C$, then we have $\nins \ge K$ and $2K\nins \le n \le \nins \cdot \frac{\epsilon^2 \sqrt{d/\nins}}{16 \log d}$. 

\begin{lemma}[\textbf{Part (1)}]
  In the inductive threat model,
  the learned model $\widetilde{F} = \Gamma(\mathcal{T}(D), D, U)$,
  by {\em any} algorithms $\mathcal{T}$ and $\Gamma$,
  must have a large error:
  \begin{align}
    \mathbb{E} \left\{ L(\widetilde{F}, \widetilde{V'}) \right\}
    \ge \frac{1}{2} (1-d^{-1}) \ge  \frac{1}{2} (1- \nu),
  \end{align}
  where the expectation is over the randomness of $D, V$ and possible algorithm randomness.
\end{lemma}

\begin{proof}
  This follows from Corollary 23 in~\cite{schmidt2018adversarially}.
  The only difference of our setting from theirs is that
  we additionally have unlabeled data $U$ for the algorithm.
  Since the attacker can provide $x'={x}$, the problem reduces to a problem
  with at most $n+1$ data points in their setting, and thus the statement follows.
\end{proof}

\noindent\textbf{Transductive learning algorithms $(\calT, \Gamma)$}:
To prove the statement (2), we give concrete learning algorithms that achieves
small test error on $x'$. We consider learning a linear classifier $\mathrm{sign}(\theta^\top x)$ with a parameter vector $\theta$. 

\vskip 5pt
\noindent\textbf{High-level structure of the learning algorithms}.
At the high level, the learning algorithms work as follows:
At the training time we use part of the training data (denoted as $D_2$ to train
a pretrained model $\bar\theta$), and part of the training data (denoted as $D_1$,
is reserved to test-time adaptation). Then, at the test time, upon receiving $U$,
we use $U$ to tune $\bar\theta$, and get two large-margin classifiers,
$\bar\theta_+$ and $\bar\theta_-$, which classify $x'$ as $+1$ and $-1$, respectively.
Finally, we check these two large margin classifiers on $D_1$ (that's where $D_1$
is used), and the one that generates smaller error wins and we classify $x'$ into
the winner class.
\vskip 5pt
\noindent\textbf{Detailed description.}
More specifically, the learning algorithms $(\calT, \Gamma)$ work as follows:
\begin{enumerate}
\item \textbf{Before game starts}.
  Let $m' = K\nins, m = 10 \nins$. We split the training set $D$ into two subsets:
  $D_1 := \{(x_i,y_i)\}_{i=1}^{m'}$ and
  $D_2 := \{(x_{m'+i},y_{m'+i})\}_{i=1}^{m}$.
  $D_2$ will be used to train a pretrained model at the training time,
  and $D_1$ will be used at the test time for adaptation.
\item \textbf{Training time}. $\mathcal{T}$ uses the second part $D_2$
  to compute a pretrained model, that is, a parameter vector:
  \begin{align}
    \hat\theta_m = \frac{1}{m} \sum_{i=1}^m y_{m'+i}x_{m'+i},\quad
    \bar\theta = \frac{\hat\theta_m}{\|\hat{\theta}_m\|_2}.
  \end{align}
\item \textbf{Test time}. On input $U$$, \Gamma$ uses $D_1$ and $U$ to perform adaptation.
  At the high level, it adapts the pre-trained $\bar{\theta}$
  along the direction of $x'$, such that it also has a large margin on $x'$,
  and also it makes correct predictions on $D_1$ with large margins. More specifically:
  \begin{enumerate}
  \item First, $\Gamma$ constructs two classifiers, $\theta_+$ and $\theta_-$,
    such that $\theta_+$ classifies $x'$ to be $+1$ with a large margin,
    and $\theta_-$ classifies $x'$ to be $-1$ with a large margin. Specifically:
    \begin{align}
      \bar{x'} & := x' / \|x'\|_2,
      & \gamma  := \|x'\|_2 / 2, \\
      \eta_+ & :=  \frac{\gamma - (\bar{\theta})^\top x'}{\|x'\|_2},
      & \theta_+ = \bar\theta + \eta_+ \bar{x'},
             & & \bar\theta_+ = \theta_+/\|\theta_+\|_2, \\
      \eta_{-} & :=  \frac{- \gamma - (\bar{\theta})^\top x'}{\|x'\|_2},
      &  \theta_- = \bar\theta + \eta_- \bar{x'},
             & & \bar\theta_- = \theta_-/\|\theta_-\|_2.
    \end{align}
    where $\theta_+$ and $\theta_-$ are viewed as the parameter vectors
    for linear classifiers. Note that $\theta_+$ is constructed such that
    $\theta_+^\top x'/\|x'\|_2 = \gamma/\|x'\|_2 = 1/2$, and $\theta_-$ is such that $\theta_-^\top x'/\|x'\|_2 = -\gamma/\|x'\|_2 = -1/2$.
  \item Finally, $\Gamma$ checks their large margin errors on $D_1$. Formally, let
    \begin{align}
      t & := \sigma\left(\sqrt{\frac{\nins}{d}} + \frac{\nins}{m} \right)^{-1/2}, \\
      \errt{t}(\theta) & := \mathbb{E}_{(x,y)} \mathbb{I}[y \theta^\top x \le t], \\
      \widehat{\errt{t}}(\theta) & := \frac{1}{m'} \sum_{i=1}^{m'}
                                   \mathbb{I}[y_i \theta^\top x_i \le t].
    \end{align}
    If $\widehat{\errt{t}}(\bar\theta_+) \le \widehat{\errt{t}}(\bar\theta_-)$,
    then $\Gamma$ sets $\widetilde{F}(x) := \sign(\bar\theta_+^\top x)$ and classifies $x'$
    to $+1$; otherwise, it sets $\widetilde{F}(x) := \sign(\bar\theta_-^\top x)$
    and classifies $x'$ to $-1$.
  \end{enumerate}
\end{enumerate}

\begin{lemma}[\textbf{Part (2)}]
  In the transductive threat model, for the $\calT$ and $\Gamma$ described above,
  the adapted model $\widetilde{F} = \Gamma(\mathcal{T}(D), D, U)$ has a small error:
  \begin{align}
    \mathbb{E} \left\{ L(\widetilde{F}, \widetilde{V}) \right\} \le \nu.
  \end{align}
\end{lemma}
\begin{proof}
  Now, we have specified the algorithms and are ready to prove that w.h.p.
  $\widetilde{F}(x')$ is the correct label $y$. 
  By Lemma~\ref{lem:errt}, $y(\errt{t}(\bar\theta_-) - \errt{t}(\bar\theta_+)) \ge \frac{c_4}{\sqrt{\nins}}$ with probability $\ge 1 - e^{-c_4 K}$. Then by the Hoeffding's inequality, $D_1$ is sufficiently large to ensure
  $y (\widehat{\errt{t}}(\bar\theta_+) - \widehat{\errt{t}}(\bar\theta_-)) > 0$
  with probability $\ge 1 - 2 e^{-c_4^2 K/2}$. This gives
  \begin{align}
    \mathbb{E} \left\{ L(\widetilde{F}, \widetilde{V}) \right\} \le e^{-cK}.
  \end{align}
  This is bounded by $\nu$ for the choice of $K$.
\end{proof}

\textbf{Tools.} We collect a few technical lemma below.
\begin{lemma} \label{lem:errt}
  There exists absolute constants $c_4 >0 $ such that with probability
  $\ge 1 - e^{-c_4 K}$,
  \begin{align}
    y(\errt{t}(\bar\theta_-) - \errt{t}(\bar\theta_+)) \ge \frac{c_4}{\sqrt{\nins}}.
  \end{align}
\end{lemma}
\begin{proof}
  Without loss of generality, assume $y = +1$. The proof for $y=-1$ follows the same argument. 

  Note that
  \begin{align}
    \errt{t}(\theta) & = \mathbb{E}_{(x,y)} \mathbb{I}[y \theta^\top x \le t] 
    \\
                     & = \mathbb{P}\left( \mathcal{N}(\mu^\top \theta, \sigma^2 \|\theta\|^2_2) \le t \right) 
    \\
                     & = Q\left( \frac{\mu^\top \theta -t}{\sigma \|\theta\|_2} \right),
  \end{align}
  where 
  \begin{align}
    Q(x) := \frac{1}{\sqrt{2\pi}} \int_x^{+\infty} e^{-t^2/2} dt.
  \end{align}

  First, consider $\bar\theta$.
  \begin{align}
    \errt{t}(\bar\theta) 
    & = Q\left( \frac{\mu^\top \bar\theta -t}{\sigma \|\bar\theta\|_2} \right) = Q\left( s \right), \textrm{~where~} s := \frac{\mu^\top \bar\theta -t}{\sigma \|\bar\theta\|_2} = \frac{\mu^\top \bar\theta -t}{\sigma }.
  \end{align}
  By Lemma~\ref{lem:inner_au}, we have with probability $ \ge 1 - e^{-c_2 (d/\nins)^{1/4} \min\{m, (d/\nins)^{1/4}\}}$,
  \begin{align}
    \frac{\mu^\top \bar\theta}{\sigma \|\bar\theta\|_2} & \le  \left(\sqrt{\frac{\nins}{d} } + \frac{n_0}{m}\right )^{-1/2}  \left( 1 + c_1 \left(\frac{\nins}{d}\right)^{1/8} \right), \\
    \frac{\mu^\top \bar\theta}{\sigma \|\bar\theta\|_2}
                                                        & \ge
                                                          \left(\sqrt{\frac{\nins}{d} } + \frac{n_0}{m} \right)^{-1/2} \left( 1- c_1 \left(\frac{\nins}{d}\right)^{1/8} \right),
  \end{align}
  which gives
  \begin{align}
    s & \le  c_1 \left(\frac{\nins}{d}\right)^{1/8} \left(\sqrt{\frac{\nins}{d} } + \frac{n_0}{m}\right )^{-1/2}, \\
    s & \ge - c_1 \left(\frac{\nins}{d}\right)^{1/8} \left(\sqrt{\frac{\nins}{d} } + \frac{n_0}{m}\right )^{-1/2}. 
  \end{align}
  Since $m=10 n_0$ and $d \gg n_0$, we have 
  \begin{align}
    |s| = \left| \frac{\mu^\top \bar\theta - t}{\sigma \|\bar\theta\|_2} \right|   \le 1.
  \end{align}

  Next, we have 
  \begin{align}
    \errt{t}(\theta_+) 
    & = Q\left( \frac{\mu^\top \bar\theta_+ -t}{\sigma \|\bar\theta_+\|_2} \right) = Q\left( s_+ \right), \textrm{~where~} s_+ :=  \frac{\mu^\top \bar\theta_+ -t}{\sigma \|\bar\theta_+\|_2} = \frac{\mu^\top \bar\theta_+ -t}{\sigma},
    \\
    \errt{t}(\theta_-) 
    & = Q\left( \frac{\mu^\top \bar\theta_- -t}{\sigma \|\bar\theta_-\|_2} \right) = Q\left( s_- \right), \textrm{~where~} s_- :=  \frac{\mu^\top \bar\theta_- -t}{\sigma \|\bar\theta_-\|_2} =  \frac{\mu^\top \bar\theta_- -t}{\sigma }.
  \end{align}
  We now check the sizes of $s_+$ and $s_-$. 
  \begin{align}
    s_+ - s & = \frac{\mu^\top \bar\theta_+ -t}{\sigma } - \frac{\mu^\top \bar\theta -t}{\sigma }
    \\
            & = \frac{\mu^\top \bar\theta_+ - \mu^\top \bar\theta}{\sigma } 
    \\
            & = \frac{1}{\sigma \|\theta_+\|_2} \big( (1-\|\theta_+\|_2) \mu^\top \bar\theta  + \eta_+ \mu^\top \bar{x'} \big).
  \end{align}
  Then by definition and bounds in Claim~\ref{claim:inners},
  \begin{align}
    |s_+ - s| \le \frac{2}{\nins} + 40 \le 42.
  \end{align}
  Since $|s|$ is bounded by 1, we know $|s_+|$ is also bounded by 43.
  Similarly, $|s_- - s|$ and thus $|s_-|$ are also bounded by some constants. 
  Furthermore,
  \begin{align}
    s_+ - s_- & = \frac{1}{\sigma}\left( \mu^\top \bar{\theta}_+ - \mu^\top \bar{\theta}_- \right) 
    \\
              & = \frac{1}{\sigma}\left( \frac{\mu^\top \bar{\theta} + \eta_+ \mu^\top \bar{x'}}{\|\theta_+\|_2} - \frac{\mu^\top \bar{\theta} + \eta_- \mu^\top \bar{x'}}{\|\theta_-\|_2} \right).
  \end{align}
  By Claim~\ref{claim:eqnorm}, we have $\|\theta_-\|_2 = \|\theta_+\|_2$. Together with bounds in Claim~\ref{claim:inners}, we have
  \begin{align}
    s_+ - s_-
    & = \frac{1}{\sigma \|\theta_+\|_2}\left( \eta_+ \mu^\top \bar{x'} -  \eta_- \mu^\top \bar{x'} \right)
    \\
    & = \frac{1}{\sigma \|\theta_+\|_2}\left( \eta_+  -  \eta_-\right)  \mu^\top \bar{x'} 
    \\
    & = \frac{1}{\sigma \|\theta_+\|_2} \mu^\top \bar{x'} 
    \\
    & \ge \frac{\sqrt{d}}{4 \sigma^2} 
    \\
    & = \frac{1}{4\sqrt{\nins}}.
  \end{align}
  Now we are ready to bound the error difference:
  \begin{align}
    \errt{t}(\bar\theta_-) - \errt{t}(\bar\theta_+) 
    & = Q(s_-) - Q(s_+)
    \\
    & = \frac{1}{\sqrt{2\pi}} \int_{s_-}^{s_+} e^{-t^2/2} dt
    \\
    & \ge \frac{1}{\sqrt{2\pi}}  (s_- - s_+) \times \min\{e^{-s_-^2/2}, e^{-s_+^2/2}\}
    \\
    & \ge \frac{c_4}{\sqrt{\nins}}
  \end{align}
  for some absolute constant $c_4 > 0$.
\end{proof}

\begin{claim} \label{claim:inners}
  There exists a absolute constant $c_3>0$,
  such that with probability $\ge 1 - e^{-c_3 K}$,
  \begin{align}
    \sigma \sqrt{d}/4 & \le \|x'\|_2  \le 2 \sigma \sqrt{d},
    \\
    \frac{1}{2}\sigma \le \frac{1}{4}\sigma \sqrt{\frac{m}{\nins}} & \le \bar\theta^\top \mu  \le 2\sigma \sqrt{\frac{m}{\nins}} \le 10 \sigma,
    \\
    -\epsilon \sqrt{d}/2 & \le \bar\theta^\top {x'}   \le 2 \epsilon \sqrt{d},
    \\
    d/2 & \le \mu^\top x' \le 3d/2,
    \\
    \frac{1}{2} - \frac{8\epsilon}{\sigma} & \le \eta_+  \le \frac{1}{2} +  \frac{8\epsilon}{\sigma},
    \\
    -\frac{1}{2} - \frac{8 \epsilon}{\sigma} & \le \eta_-  \le - \frac{1}{2} +  \frac{8\epsilon}{\sigma}.
  \end{align}
\end{claim}
\begin{proof}
  First, since $x' = \mu + \sigma \zeta + \nu$ for $\zeta \sim \mathcal{N}(0,I)$, with probability $\ge 1 - e^{-c' d}$ for an absolute constant $c'>0$, we have:
  \begin{align}
    \sqrt{d}/2 & \le \|\zeta\|_2  \le 3\sqrt{d}/2,
    \\
    \|x'\|_2 & \ge \sigma \sqrt{d}/2 - \|\mu\|_2 - \|\nu\|_2 \ge \sigma \sqrt{d}/4,
    \\
    \|x'\|_2 & \le \sigma 3\sqrt{d}/2 + \|\mu\|_2 + \|\nu\|_2 \le  2\sigma \sqrt{d}. 
  \end{align}
  By Lemma~\ref{lem:inner_au}, with probability $ \ge 1 - e^{-c_2 K}$,
  \begin{align}
    \bar\theta^\top \mu & \le 2\sigma \left( \sqrt{\frac{\nins}{d} }  + \frac{\nins}{m}\right)^{-1/2} \le 2\sigma \sqrt{\frac{m}{\nins}},
    \\
    \bar\theta^\top \mu & \ge \frac{1}{2}\sigma \left( \sqrt{\frac{\nins}{d} }  + \frac{\nins}{m}\right)^{-1/2} \ge \frac{\sigma}{4} \sqrt{\frac{m}{\nins}}.
  \end{align}
  Also, with probability $1 - e^{-c'K}$,
  \begin{align}
    |\bar\theta^\top \zeta| \le 2 K \sigma.
  \end{align}
  Finally,
  \begin{align}
    |\bar\theta^\top \nu| \le \|\bar\theta\|_1 \|\nu\|_\infty \le \epsilon\sqrt{d}.
  \end{align}
  Then
  \begin{align}
    \bar\theta^\top {x'} & = \bar\theta^\top (\mu + \sigma \zeta + \nu) 
    \\
                         & \le |\bar\theta^\top \mu| + \sigma |\bar\theta^\top   \zeta| + |\bar\theta^\top \nu|
    \\
                         & \le  2 \sigma \sqrt{\frac{m}{\nins}} + 2 K \sigma + \epsilon \sqrt{d}
    \\
                         & \le 2 \epsilon \sqrt{d}. 
  \end{align}
  and
  \begin{align}
    \bar\theta^\top {x'} & = \bar\theta^\top (\mu + \sigma \zeta + \nu)
    \\
                         & \ge \sigma/2 - K \sigma  - \epsilon \sqrt{d}
    \\ 
                         & \ge - \epsilon \sqrt{d}/2. 
  \end{align}
  For $\mu^\top x'$, we have with probability $\ge 1 - e^{-c'K}$,
  \begin{align}
    \mu^\top x'  &  =  \mu^\top (\mu + \sigma \zeta + \nu)
    \\
    \mu^\top x'  &  \le \|\mu\|_2^2 + 2 K \sigma \|\mu\|_2 + \epsilon \|\mu\|_2 \sqrt{d} \le 3 d/ 2,
    \\
    \mu^\top x'  &  \ge \|\mu\|_2^2 - 2 K \sigma \|\mu\|_2 - \epsilon \|\mu\|_2 \sqrt{d} \ge d/2.
  \end{align}
  By definition:
  \begin{align}
    \eta_+ & = \frac{1}{2} - \bar{\theta}^\top x' /\|x'\|_2,
  \end{align}
  so
  \begin{align}
    \frac{1}{2} - 8\epsilon / \sigma \le \eta_+ & \le \frac{1}{2} + 8\epsilon / \sigma.
  \end{align}
  Similarly,
  \begin{align}
    -\frac{1}{2} - 8\epsilon / \sigma \le \eta_- & \le - \frac{1}{2} + 8\epsilon / \sigma.
  \end{align}
\end{proof}

\begin{claim} \label{claim:eqnorm}
  \begin{align}
    \|\theta_+\|_2  = \|\theta_-\|_2.
  \end{align}
\end{claim}
\begin{proof}
  We have by definition:
  \begin{align}
    \|\theta_-\|^2_2 
    & = \|\bar\theta + \eta_- \bar{x'} \|^2_2 
    \\
    & =  1 + \eta_-^2 + 2 \eta_- \bar\theta^\top \bar{x'},
    \\
    \|\theta_+\|^2_2 
    & = \|\bar\theta + \eta_+ \bar{x'} \|^2_2 
    \\
    & =  1 + \eta_+^2 + 2 \eta_+ \bar\theta^\top \bar{x'}.
  \end{align}
  Then
  \begin{align}
    \|\theta_-\|^2_2  - \|\theta_+\|^2_2 
    & =  \eta_-^2 + 2 \eta_- \bar\theta^\top \bar{x'} - \eta_+^2 - 2 \eta_+ \bar\theta^\top \bar{x'}
    \\
    & = (\eta_- - \eta_+)(\eta_- + \eta_+) + 2 \bar\theta^\top \bar{x'} (\eta_- - \eta_+)
    \\
    & = (\eta_- - \eta_+)[(\eta_- + \eta_+) + 2 \bar\theta^\top \bar{x'} ]
    \\
    & = (\eta_- - \eta_+)[-2 \bar\theta^\top x'/\|x'\|_2 + 2 \bar\theta^\top \bar{x'} ]
    \\
    & = 0.
  \end{align}
  This completes the proof.
\end{proof}

\begin{lemma}[Paraphrase of Lemma 1 in~\cite{carmon2019unlabeled}]\label{lem:inner_au}
  Let $\hat{\theta}_m = \frac{1}{m} \sum_{i=1}^m y_i x_i$.
  There exist absolute constants $c_0, c_1, c_2$ such that for $\|\mu\|^2_2 = d,   \sigma^2 = \sqrt{d \nins}$, and $d/\nins > c_0$, 
  \begin{align}
    \frac{\sigma^2  \|\hat\theta_{m}\|^2_2}{(\mu^\top \hat\theta_{m})^2} 
    & \ge \left(\sqrt{\frac{\nins}{d} } + \frac{n_0}{m}\right )  \left( 1 - c_1 \left(\frac{\nins}{d}\right)^{1/8} \right), \\
    \frac{\sigma^2 \|\hat\theta_{m}\|^2_2}{(\mu^\top \hat\theta_{m})^2} 
    & \le
      \left(\sqrt{\frac{\nins}{d} } + \frac{n_0}{m} \right) \left( 1+ c_1 \left(\frac{\nins}{d}\right)^{1/8} \right),
  \end{align}
  with probability $ \ge 1 - e^{-c_2 (d/\nins)^{1/4} \min\{m, (d/\nins)^{1/4}\}}$.
\end{lemma}
